\documentclass[journal]{IEEEtran}
\usepackage{xcolor}
\usepackage{url}
\usepackage{ifpdf}
\usepackage{cite}
\usepackage[pdftex]{graphicx}
\usepackage{amsmath}
\usepackage{algorithmic}
\usepackage{framed,multirow}
\usepackage{siunitx}
\usepackage{textcomp}
\usepackage{gensymb}
\usepackage{amssymb}
\usepackage{subfigure}
\usepackage{xspace}
\allowdisplaybreaks
\usepackage[colorlinks,linkcolor=red]{hyperref}
\usepackage{tikz}
\newcommand*{\circled}[1]{\lower.7ex\hbox{\tikz\draw (0pt, 0pt)%
circle (.43em) node {\makebox[0em][c]{\small #1}};}}

\ifCLASSINFOpdf
\else
\fi
\begin{document}
\title{6D-ViT: Category-Level 6D Object Pose Estimation via Transformer-based Instance Representation Learning}
\author{Lu~Zou,~\IEEEmembership{Student~Member,~IEEE,}
        Zhangjin~Huang,~\IEEEmembership{Member,~IEEE,}
        Naijie~Gu, 
        and Guoping~Wang%,~\IEEEmembership{Member,~IEEE}%
%\author{Lu~Zou,~\IEEEmembership{Student~Member,~IEEE,}
%       Zhangjin~Huang$^\dagger$
% <-this % stops a space
\thanks{Lu Zou, Zhangjin Huang, and Naijie Gu are with University of Science and Technology of China, Hefei 230031, China (e-mail: lzou@mail.ustc.edu.cn; zhuang@ustc.edu.cn; gunj@ustc.edu.cn). (Corresponding author: Zhangjin Huang)}% <-this % stops a space
\thanks{Guoping Wang is with Peking University, Beijing 100871, China (e-mail: wgp@pku.edu.cn).}% <-this % stops a space

%\author{Michael~Shell,~\IEEEmembership{Member,~IEEE,}
%John~Doe,~\IEEEmembership{Fellow,~OSA,} 
%and~Jane~Doe,~\IEEEmembership{Life~Fellow,~IEEE} % <-this % stops a space
%\thanks{M. Shell was with the Department
%of Electrical and Computer Engineering, Georgia Institute of Technology, Atlanta,
%\thanks{$^\dagger$Corresponding author.}
\thanks{Manuscript received April 19, 2005; revised August 26, 2015. }} %Editor: Please ensure that all of the text that is serving as a placeholder is appropriately replaced or removed.


\markboth{Journal of \LaTeX\ Class Files,~Vol.~14, No.~8, August~2015}%
{Shell \MakeLowercase{\textit{et al.}}: Bare Demo of IEEEtran.cls for IEEE Journals}



	\maketitle

\begin{abstract}
This paper presents 6D vision transformer (6D-ViT), a transformer-based instance representation learning network that is suitable for highly accurate category-level object pose estimation on RGB-D images. Specifically, a novel two-stream encoder-decoder framework is dedicated to exploring complex and powerful instance representations from RGB images, point clouds and categorical shape priors. The whole framework consists of two main branches, named Pixelformer and Pointformer. Pixelformer contains a pyramid transformer encoder with an all-multilayer perceptron (MLP) decoder to extract pixelwise appearance representations from RGB images, while Pointformer relies on a cascaded transformer encoder and an all-MLP decoder to acquire the pointwise geometric characteristics from point clouds. Then, dense instance representations (\textit{i.e}., correspondence matrix and deformation field) are obtained from a multisource aggregation (MSA) network with shape prior, appearance and geometric information as input. Finally, the instance 6D pose is computed by leveraging the correspondence among dense representations, shape priors, and instance point clouds. Extensive experiments on both synthetic and real-world datasets demonstrate that the proposed 3D instance representation learning framework achieves state-of-the-art performance on both types of datasets and significantly outperforms all existing methods. Our code will be available. %Code is available at \url{https://github.com/luzou-ustc/6D-ViT}.
\end{abstract}

\begin{IEEEkeywords}
6D object pose estimation, 3D object detection, vision transformer, representation learning.
\end{IEEEkeywords}

% \fi
\IEEEpeerreviewmaketitle

\section{Introduction}
\IEEEPARstart{O}{bject} pose estimation refers to predicting the location and orientation of 3D objects, which is a fundamental problem in robotic applications such as grasping and manipulation. In recent years, remarkable progress~\cite{hinterstoisser2012model,kehl2016deep,kehl2017ssd,krull2015learning,li2018unified,michel2017global,peng2019pvnet,sundermeyer2018implicit,tekin2018real,wang2019densefusion,xiang2017posecnn,zakharov2019dpod} has been made on instance-level 6D object pose estimation, where exact 3D CAD models and their sizes are available in advance. Unfortunately, considering the diversity of object instances and the cost of building a CAD model for each instance, these methods are difficult to generalize to real practice.

\begin{figure}[!ht]
		\centering %
		\includegraphics[scale=0.55]{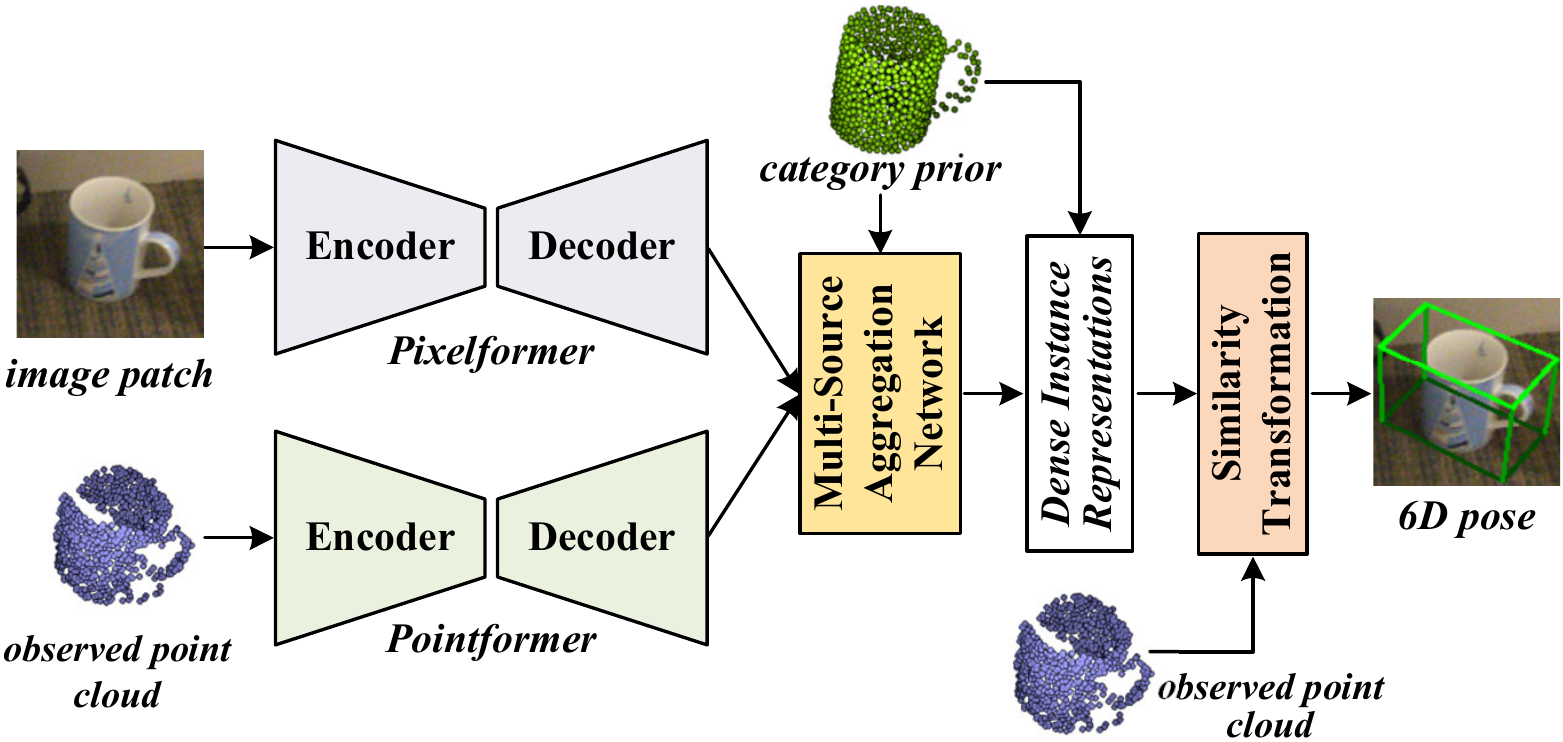}
		\label{fig:framework}
\caption{The main network of our transformer-based instance representation learning network for category-level 6D object pose estimation.}
\end{figure}

%these methods cannot be directly generalized to category-level 6D object pose estimation where the vast majority of the objects have never been seen before and have no known CAD models for the instances.

Recently, category-level 6D object pose estimation has begun to receive increasing attention~\cite{chen2020category,chen2020learning,chen2021fs,lin2021dualposenet,tian2020shape,wang2019normalized} given its practical importance. Compared with the instance-level problem, the goal of this task is to predict the 6D pose of unseen object instances of the same category for which no CAD models are available. In general, the classical instance-level pose estimation methods usually predict object poses by establishing the correspondence between an RGB or RGB-D image and the CAD model of the object. %Editor: Please ensure that the intended meaning has been maintained in this edit.
However, in the category-level scenario, since there are no specific CAD models, it is quite difficult to directly establish this correspondence. In addition, even within the same category, objects can exhibit significant variations in shape, color, and size. Consequently, due to the large intraclass variations in texture and shape among instances, category-level pose estimation is much more challenging than instance-level problems~\cite{sahin2019instance}.

To achieve reliable category-level object pose estimation, Wang \textit{et al}.~\cite{wang2019normalized} innovatively proposed normalized object coordinate space (NOCS) as a shared canonical representation of all object instances within a category. Since the CAD model of the object is unknown, they reconstruct the canonical model representation in the NOCS and establish the dense correspondence between the reconstructed NOCS model and instance images or point clouds to realize 6D pose estimation. Later, Tian \textit{et al.}~\cite{tian2020shape} improved the NOCS model reconstruction process by introducing shape priors. They extract category features from the shape priors as auxiliary information of instance features to alleviate the shape variations across instances within the same category. Recently, Chen \textit{et al}.~\cite{chen2021fs} proposed the fast shape-based network (FS-Net) to further improve the quality of category-level pose features. They employ a shape-based 3D graph convolution autoencoder to extract the latent features through the observed point reconstruction and propose a decoupled rotation mechanism to decode instance rotation information from the latent features. Despite the promising performance gains, this method is too complex and requires training separate models for different object categories due to their specific data augmentation.

In this work, we tackle the problem of category-level object pose estimation from the perspective of unified and powerful instance representation learning on RGB-D data. More specifically, inspired by the most famous transformer architectures~\cite{liu2021transformer,vaswani2017attention,wang2021pyramid,xie2021segformer,zhao2021pointtransformer} widely considered by the natural language processing and computer vision communities, we develop a two-stream transformer-based encoder-decoder architecture, 6D vision transformer (6D-ViT), for highly accurate category-level object pose estimation. Our network is composed of the Pixelformer and Pointformer modules, which are used to explore the appearance and geometric characteristics of the object instance, respectively. Then, the dense representations (\textit{i.e}., correspondence matrix and deformation field) of the object instance are obtained through the fusion of appearance information and geometric information from different sources. Finally, the category-level 6D pose is obtained by establishing the correspondence among dense representations, categorical shape priors, and the observed point clouds. Comprehensive experiments on both synthetic (CAMERA25) and real-world (REAL275) datasets suggest that the proposed method substantially improves in terms of performance on both datasets and achieves state-of-the-art performance through a unified model for all categories.

Our contributions can be summarized as follows.

\begin{itemize}
\item We present a novel two-stream category-level 6D object pose estimation network based on transformer architectures. The network can semantically enhance the quality of instance representations by capturing the long-range contextual dependencies of elements from RGB images and point clouds.
\item Our network includes Pixelformer and Pointformer, both of which are encoder-decoder architectures for dense instance representation learning. In particular, both encoders are multistage transformer architectures for complex feature extraction, while both decoders are all-MLP architectures for efficient feature aggregation.
\item Extensive experiments demonstrate that our proposed 6D-ViT achieves state-of-the-art performance on both the CAMERA25 and REAL275 datasets and significantly outperforms existing works.
\end{itemize}

The remainder of this paper is organized as follows. Section~\ref{sec:related_work} briefly reviews some related works about 6D object pose estimation including instance-level (Section~\ref{sec:instance_level}) and category-level (Section~\ref{sec:category_level}) works. Section~\ref{sec:method} introduces the presented method in detail, including the overall framework (Section~\ref{sec:overview}), instance representation learning on RGB images (Section~\ref{sec:rgbformer}), instance representation learning on point clouds (Section~\ref{sec:pointformer}), joint and dense representation generation and pose estimation (Section~\ref{sec:dense_fusion}) and loss functions (Section~\ref{sec:loss}). The experimental results of our network with comparisons to existing methods are reported in Section~\ref{sec:experiments}. Specifically, we evaluate the proposed method in terms of 6D object pose estimation (Section~\ref{sec:eva_pose}) and 3D model reconstruction (Section~\ref{sec:eva_recons}). In addition, the ablation studies of our network are discussed in Section~\ref{sec:ablation}. Finally, conclusions and future directions are provided in Section~\ref{sec:conclusion}.

\section{Related Work}
	\label{sec:related_work}

\subsection{Instance-Level 6D Object Pose Estimation}
	\label{sec:instance_level}

A large amount of research has focused on instance-level object pose estimation. In this section, we discuss only some of the most notable studies. According to the format of the input data, instance-level pose estimation methods can be broadly divided into RGB-based and RGB-D-based methods. Traditional RGB-based methods~\cite{hinterstoisser2012model,zhu2014single} realize pose estimation by detecting and matching keypoints
with the known 3D CAD model of the object. However, these methods rely heavily on handcrafted features that are not robust to low-texture objects and cluttered environments. Recently, many works have applied deep learning techniques to this task due to their robustness to environmental variations. Kehl \textit{et al}.~\cite{kehl2017ssd} extended the 2D object detection network~\cite{liu2016ssd} to predict the identity of an object, the 2D bounding box and the discretized orientation. Tekin \textit{et al}.~\cite{tekin2018real} proposed first detecting the keypoints of the object and then solving a perspective-n-point problem for pose estimation. Zakharov \textit{et al}.~\cite{zakharov2019dpod} estimated the dense 2D-3D correspondence map between the input image and the object model. Peng \textit{et al}.~\cite{peng2019pvnet} presented the prediction of a unit vector for each pixel pointing toward the keypoints. %Song \textit{et al}.~\cite{song2020hybridpose} improved~\cite{peng2019pvnet} by utilizing keypoints, edge vectors, and symmetry correspondence to express different geometric information in the input image. 

As an increasing number of RGB-D datasets are available, another line of work~\cite{kehl2016deep,li2018unified,wang2019densefusion} began to apply both RGB and depth images to improve the pose estimation accuracy. Considering the different distributions of RGB and depth data, there are several different ways to deal with these two modalities. Kehl \textit{et al}.~\cite{kehl2016deep} simply concatenated RGB and depth values and fed the 4-channel data into a multistage model to produce pose estimations. Xiang \textit{et al}.~\cite{xiang2017posecnn} first predicted a rough 6D pose from an RGB image, followed by the application of the iterative closest point (ICP) algorithm~\cite{besl1992method} using depth images for refinement. Wang \textit{et al}.~\cite{wang2019densefusion} proposed DenseFusion; they design a two-branch framework to exploit the features of RGB and depth images separately and fuse the features of different modalities through a fully connected (FC) layer. DenseFusion achieves state-of-the-art performance while reaching almost real-time inference speed. Nevertheless, it merges the features of two modalities in a straightforward manner, which is not sufficient to capture their inherent correlations. To this end, Zou \textit{et al}.~\cite{zou2021cma} further improved the feature fusion process in DenseFusion~\cite{wang2019densefusion} and developed a cross-modality attention scheme to learn the feature interactions between the RGB images and point clouds. Our approach also estimates 6D object poses from RGB-D images; however, we focus on a more general setting where object CAD models are not available during inference.

\subsection{Category-Level 6D Object Pose Estimation}
	\label{sec:category_level}

Few previous works~\cite{wang2019normalized,lin2021dualposenet,chen2020learning,chen2020category,tian2020shape,chen2021fs} have focused on estimating the 6D poses of unseen objects. Compared to instance-level problems, category-level tasks are much more challenging due to the large intraclass variations in the aspects of texture and shape among instances. To address the above issue, some researchers have proposed constructing an intermediate instance representation to mitigate such differences. Wang \textit{et al}.~\cite{wang2019normalized} represented different object instances within a category as a shared NOCS. They train a region-based deep network to infer the correspondence from observed pixels to points in the NOCS. As a result, the 6D pose and size can be calculated by shape matching between the predicted correspondence and the observed points. Chen \textit{et al}.~\cite{chen2020learning} introduced a unified category representation for a large variety of instances called canonical shape space (CASS). Additionally, a variational autoencoder (VAE) is trained to estimate the category-level object pose. Similar to~\cite{chen2020learning}, Tian \textit{et al}.~\cite{tian2020shape} proposed shape prior deformation (SPD), which leverages the category-sensitive features to explicitly model the shape variation when reconstructing the NOCS model. Specifically, they enhance the RGB-D features by introducing shape priors as an aid to the process of NOCS model reconstruction. To better explore pose-related features, Chen \textit{et al}.~\cite{chen2021fs} proposed FS-Net, which first extracts the latent instance features through the observed point reconstruction with a shape-based 3D graph convolution autoencoder. Then, a decoupled rotation mechanism is proposed to decode instance rotation information from the instance features. To increase the generalization ability of the network, an online 3D deformation mechanism is proposed for data augmentation. Despite the impressive performance gains, this method is too complex, and it requires the training of separate models for different object categories due to their specific data preprocessing, making it inconvenient for practical use. More recently, Lin \textit{et al.}~\cite{lin2021dualposenet} presented DualPoseNet to learn rotation-equivariant shape features; it stacks two parallel pose decoders at the top of a shared pose encoder, where the implicit decoder predicts object poses with a working mechanism different from that of the explicit one. Thus, complementary supervision is imposed on the training of the pose encoder.

Our work is based on SPD~\cite{tian2020shape} and takes inspiration from the latest transformer architectures~\cite{liu2021transformer,vaswani2017attention,wang2021pyramid,xie2021segformer,zhao2021pointtransformer} successfully applied in natural language processing and computer vision. Specifically, we design two transformer-based encoder-decoder networks to individually explore the compact instance representations from the RGB images and point clouds. The experimental results demonstrate that our method achieves state-of-the-art performance and significantly outperforms existing works.

\section{Method}
	\label{sec:method}

Given a calibrated RGB-D image pair, our goal is to predict the 6D object pose represented by a rigid transformation $[R|t]$, where $R \in SO(3)$ and $t \in \mathbb{R}^3$. To handle scenes containing multiple instances, we first apply an off-the-shelf instance segmentation network (\textit{i.e}., Mask R-CNN~\cite{he2017mask}) to detect and segment objects in the scene. Next, the yielded object bounding box is exploited to crop the RGB image into image patches, and the segmentation mask is leveraged to convert the depth image into observed point clouds according to the camera intrinsic parameters. After that, image patches, observed point clouds, and categorical shape priors are fed into the main part of our network.

\begin{figure*}[!ht]
		\centering
		\includegraphics[scale=0.67]{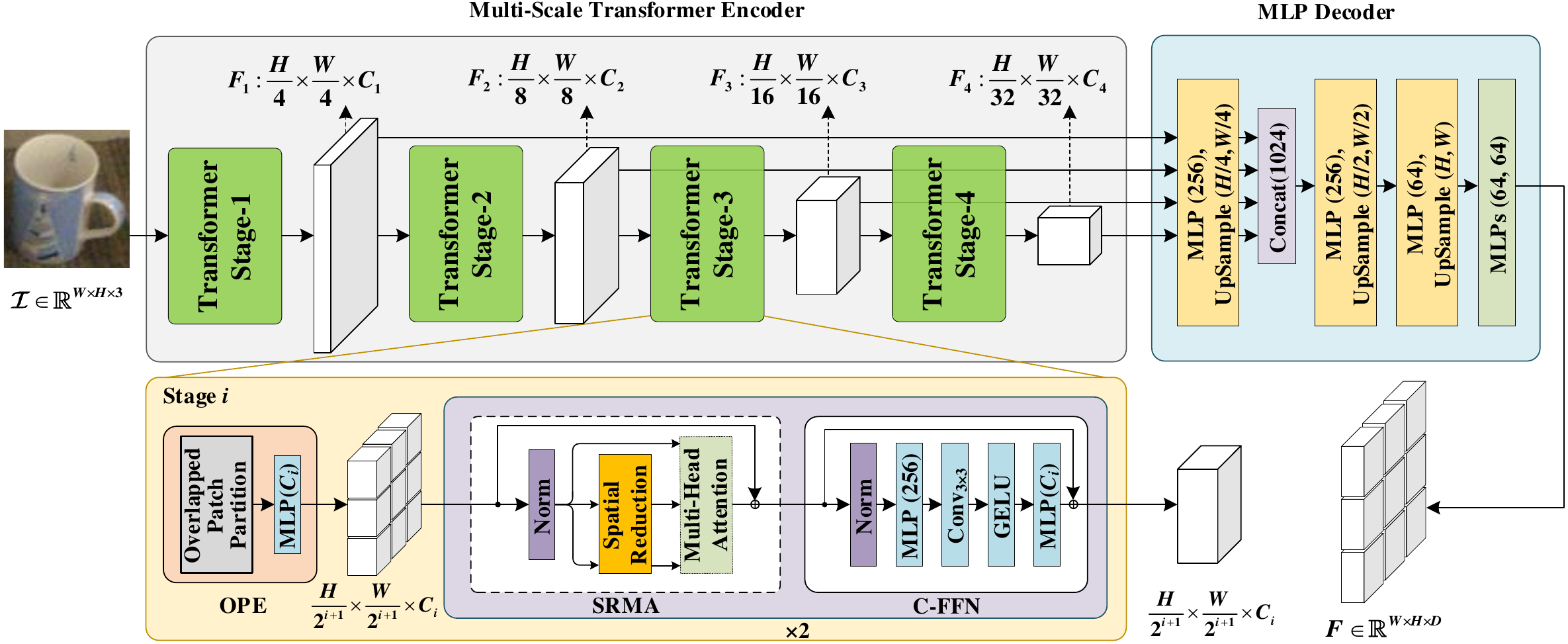}
\caption{The structure of the proposed Pixelformer. The network consists of two main modules: a multiscale transformer encoder to extract coarse and fine appearance embeddings from RGB images and an MLP decoder to fuse the multiscale embeddings and to generate the pixelwise object appearance representations. Note that in the figure, each item in the format of $Function(\cdot,...,\cdot)$ denotes the feature dimension after the corresponding operation.}
		\label{fig:fig_rgbformer}
\end{figure*}

\subsection{Architecture Overview}
	\label{sec:overview}

As shown in \figurename~\ref{fig:framework}, our main network consists of three subnetworks: (1) Pixelformer for instance representation learning on RGB images (Section~\ref{sec:rgbformer}); (2) Pointformer for instance representation learning on point clouds (Section~\ref{sec:pointformer}); and (3) a multisource aggregation (MSA) network for joint and dense representation generation and pose estimation (Section~\ref{sec:dense_fusion}). More specifically, the image patches $\mathcal{I}$ and observed point clouds $\mathcal{P}$ are first passed through Pixelformer and Pointformer to learn instance representations existing in different modalities. Then, the instance representations in the two modalities, denoted as $F$ and $\mathcal{F}$, are fused and enhanced by the category-sensitive representations extracted from the categorical shape priors $\mathcal{P}_{pri}$ to generate the joint and dense instance representations (\textit{i.e}., the correspondence matrix $\mathcal{A}$ and deformation field $\mathcal{D}_{def}$). Finally, the dense representations, categorical shape priors and observed instance point clouds are leveraged by the Umeyama algorithm~\cite{umeyama1991least} to calculate the 6D object pose $[R|t]$. In addition, the loss functions used to train the whole network are introduced in Section~\ref{sec:loss}. The design details of each step are discussed below.

\subsection{Instance Representation Learning on RGB Images}
	\label{sec:rgbformer}

As depicted in \figurename~\ref{fig:fig_rgbformer}, Pixelformer is an encoder-decoder architecture that consists of two main components: 1) a multiscale transformer encoder that contains $I$ stages to generate high-resolution coarse appearance embeddings and low-resolution fine appearance embeddings; and 2) a multilayer perceptron (MLP) decoder to upsample and aggregate the pyramid features to produce the pixelwise instance appearance representations.

\subsubsection{Multiscale Transformer Encoder}

The design of the transformer encoder in our Pixelformer is partly inspired by the latest multiscale vision transformer: (PVT)~\cite{wang2021pyramid}, and is tailored and optimized for our task. Specifically, our multiscale transformer encoder consists of four stages with different latent feature dimensions. In each stage $i$, it has an overlapped patch embedding (OPE) layer and two layers of alternating spatial reduction-based multihead attention (SRMA) and a convolutional feed-forward network (C-FFN). In addition, we employ InstanceNorm (IN) before each SRMA and C-FFN module and provide residual connections around each of them.

% of $\{C_i|i= 1, 2, 3, 4\}$ = $\{32, 64, 160, 256\}$, respectively
\paragraph{Overlapped Patch Embedding}
	\label{sec:patch_embed}
In vision transformers, patch embedding is usually used to combine nonoverlapping images or feature patches. However, since the patches are nonoverlapping, this fails to preserve the local continuity around those patches. In this work, we utilize OPE as introduced in~\cite{xie2021segformer} to tokenize images. Specifically, given an input RGB image with a resolution of $H\times W \times 3$, we define $K$ as the patch window size, $S$ as the stride between two adjacent patches, and $P$ as the padding size. We first split the input image into $\frac{H}{2^{(i+1)}}$$\cdot$$\frac{W}{2^{(i+1)}}$ patches, and each patch has a size of $K_i$$\times$$K_i$$\times C_{i-1}$, where $i \in \{1,2,3,4\}$ and $C_0 = 3$. Then, the patch embedding process performs patch merging via linear projection on these patches to produce feature maps with the same dimensions as the nonoverlapping process. In the experiments, we set $K_1=7$, $S_1=4$, $P_1=3$, and $K_j=3$, $S_j=2$, $P_j=1$, where $j \in \{2,3,4\}$. As a result, we can obtain the multiscale features $\{F_i|i=1, 2, 3, 4\}$ with the shapes of \{$\frac{H}{4}\times\frac{W}{4}\times C_1$, $\frac{H}{8}\times\frac{W}{8}\times C_2$, $\frac{H}{16}\times\frac{W}{16}\times C_3$, $\frac{H}{32}\times\frac{W}{32}\times C_4$\}. Following the design rules of ResNet~\cite{he2016deep}, we use small output channels in shallow stages, \textit{i.e}., $C_i>C_{i-1}$ for $i\in \{1,2,3,4\}$.

\paragraph{Spatial Reduction-based Multihead Attention}

Considering the original multihead attention mechanism~\cite{vaswani2017attention}, given a series of vectors $Q_i$, $K_i$, $V_i$ = ${\rm MLP}(C_i, C_i)(F_i)$, $i \in \{1,2,3,4\}$ at each stage $i$, each head of $Q_i$, $K_i$, $V_i$ has the same shape of $(\frac{H}{2^{(i+1)}}$$\cdot$$\frac{W}{2^{(i+1)}})\times d_{head(i)} = (H_i \cdot W_i)\times d_{head(i)}$. Here, $\mathcal{N} = (H_i \cdot W_i)$ is the length of a sequence, and $d_{head(i)}$ refers to the dimension of each head at each stage, which is computed as $d_{head(i)}=C_i/M_i$, with $M_i$ as the number of attention heads. Note that in the format of $Function(b,c)(a)$ in the full text, $a$ represents the input feature vector, and $b$ and $c$ represent the dimensions of the input and output features, respectively. Therefore, the multihead attention operation is calculated as:

\begin{equation}
		{\rm{Attention}}(Q_i,K_i,V_i)={\rm{Softmax}}(\frac{Q_iK_i^T}{\sqrt{d_{head(i)}}})V_i, \forall i.
		\label{eq:sa}
	\end{equation}

\begin{figure*}[!ht]
		\centering
		\includegraphics[scale=0.7]{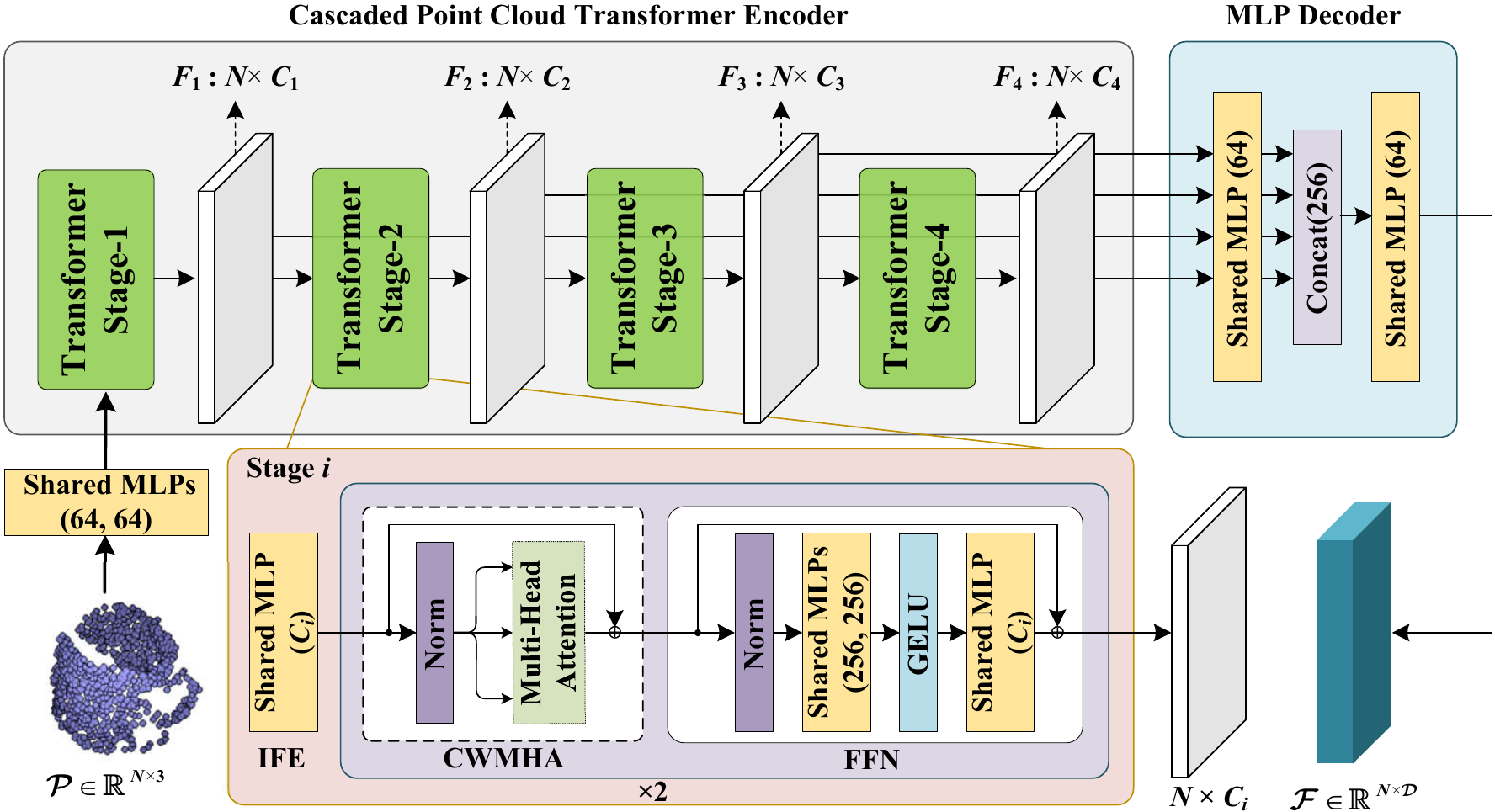}
\caption{The structure of the proposed Pointformer. The network consists of two main modules: a cascaded transformer encoder with different feature dimensions to capture multilevel point embeddings and an all-MLP decoder to unify and fuse the multilevel point embeddings and to produce the pointwise object geometric representations. Note that in the figure, each item in the format of $Function(\cdot,...,\cdot)$ denotes the feature dimension after the corresponding operation.}
		\label{fig:fig_pcformer}
\end{figure*}

The computational complexity of Eq.~\eqref{eq:sa} is $O(\mathcal{N}^2)$, which is prohibitive for large feature resolutions. Instead, we utilize the reduction process introduced in PVT~\cite{wang2021pyramid}, which uses a reduction ratio $R_i$ in stage $i$ to reduce the spatial scale of $K$ and $V$ before the attention operation executes. The spatial reduction (SR) operation is defined as follows:

\begin{equation}
		{\rm{SR}}(x)={\rm Reshape(IN}(x),R_i)\mathcal{W}_i, x \in \{K_i,V_i\}, \forall i.
		\label{eq:sr}
	\end{equation}

In Eq.~\eqref{eq:sr}, {\rm{Reshape}}($x$, $R_i$) is the operation of reshaping the input $x\in \mathbb{R}^{\mathcal{N}\times C_i}$ to the sequence with the shape of $\frac{\mathcal{N}}{R_i^2}\times(R^2_i\cdot C_i)$, and $\mathcal{W}_i \in \mathbb{R}^{(R^2_i\cdot C_i)\times C_i}$ refers to a linear projection operation that reduces the dimension of the input $x$ to $C_i$. Hence, the complexity of the self-attention mechanism is reduced from $O(\mathcal{N}^2)$ to $O(\mathcal{N}^2)/{R}$. In the experiments, we set $R^2$ to $[64,16,4,1]$ from stage one to stage four. Without abuse of notation, we denote the multiscale features after SRMA as \{$F_i|i=1,2,3,4$\}, the same as the input.

% \textcolor{blue}{By integrating the multi-scale transformer module, we can further boost the ability of our network to learn a discriminative representation for each pixel with semantically and geometrically enhanced information.}

\paragraph{Convolutional Feed-Forward Network}

Previous transformer-based image encoders employed MLPs only as their feed-forward network (FFN) and completely relied on the self-attention mechanism to achieve feature enhancement. Although effective, this design cannot perceive the locality and translation invariance of the object and ignores the interpixel dependencies that play a critical role in object pose estimation. Based on this, we propose the insertion of convolutional operations into the FFN.

Specifically, we insert a 3$\times$3 convolutional operation with a padding size of 1 between the first MLP layer and the Gaussian error linear unit (GELU)~\cite{hendrycks2016gaussian} operation. As a result, we remove the positional encoding in the traditional FFN since the convolutional operation already incorporates positional information. We formulate the C-FFN as:

\begin{align}
		\hat{F}^0_i &= {\rm MLP}(C_i, C_{expand})({\rm IN}(F_i)), \forall i, \\
		\hat{F}^1_i &= {\rm Conv_{3\times 3}}(C_{expand}, C_{expand})(\hat{F}^0_i), \forall i,\\
		\hat{F}^2_i &= {\rm GELU}(\hat{F}^1_i), \forall i, \\
		\hat{F}_i &= {\rm MLP}(C_{expand}, C_i)(\hat{F}^2_i) + F_i, \forall i,
	\end{align}

%\begin{equation}
%	\textbf{x}\_{out} = \textbf{MLP}(\textbf{GELU}(\textbf{Conv}_{3\times3}(\textbf{MLP}(\textbf{x}_{in})))) + \textbf{x}_{in},
%\end{equation}

\noindent where $C_{expand}$ represents the expanded feature dimension of the FFN~\cite{vaswani2017attention} and $\{\hat{F}_i|i= 1,2,3,4\}$ denotes the output pyramid feature maps of the C-FFN module. We set the expansion ratio of each stage to $[8, 8, 4, 4]$.

\subsubsection{Multilayer Perceptron Decoder}

We design a simple decoder to generate the unified appearance representations of the object instance from the pyramid features obtained from the multiscale transformer encoder. First, the pyramid features $\{\hat{F}_i|i=1, 2, 3, 4\}$ are passed through an MLP layer to unify the channel dimensions to be consistent with the maximum feature channel $C_4$. Then, the unified feature maps are upsampled to $1/4$ the size of the original input image and concatenated along the feature channel to form $ \frac{H}{4} \times \frac{W}{4} \times 4C_4$. Later, an MLP layer is used to fuse the concatenated features to $\frac{H}{4} \times \frac{W}{4} \times C_4$. Next, the fused features are passed through two alternative upsampling and MLP layers to obtain the unified appearance representations. We formulate the process of the MLP decoder as follows:

\begin{align}
		\hat{F}^0_i &= {\rm MLP}(C_i,C_4)(\hat{F}_i), \forall i, \\
		\hat{F}^1_i &= {\rm UpSample}(H_i \times W_i,  \frac{H}{4}\times\frac{W}{4})(\hat{F}^0_i),\forall i, \\
		\hat{F}^2   &= {\rm MLP}(4C_4,C_4)({\rm Concat}(\hat{F}^1_i)),\forall i, \\
		\hat{F}^3   &= {\rm UpSample}(\frac{H}{4}\times\frac{W}{4}, \frac{H}{2}\times\frac{W}{2})(\hat{F}^2),\\ 
		\hat{F}^4   &= {\rm MLP}(C_4,2D)(\hat{F}^3), \\ 
		\hat{F}^5   &= {\rm UpSample}(\frac{H}{2}\times\frac{W}{2}, H\times W)(\hat{F}^4), \\ 
		F         &= {\rm MLP}(2D, D)({\rm MLP}(2D,2D)(\hat{F}^5)),         
	\end{align}

\noindent where $F$ refers to the output pixelwise instance appearance representations with the shape $H\times W \times D$, where $D$ is the output dimension, which is set as $32$ to be consistent with previous appearance extractors~\cite{tian2020shape,wang2019densefusion}. Additionally, for the $UpSample(b,c)(a)$ operation, we utilize simple bilinear interpolation to recover the details of the origin feature map. The design details of Pixelformer can be found in \tablename~\ref{tab::rgbformer}.

\begin{table}[!ht]
		\centering
		\renewcommand{\multirowsetup}{\centering} 
		\caption{Detailed settings of the proposed Pixelformer for instance representation learning on RGB images.} 
		\resizebox{0.5\textwidth}{!}{\begin{tabular}{l|cccc}
				\hline
				\multirow{2}{2cm}{Param Name} &\multicolumn{4}{c}{Stage $I$}  \\\cline{2-5} 
				&1 &2 &3 &4 \\
				\hline
				Number of Attention Heads $M$ &1 &2 &5 &8\\
				FFN Expansion Ratio $R_{FFN}$ &8 &8 &4 &4\\
				Feature Channels $C$ &32 &64 &160 &256\\
				Expanded Dimension $C_{expand}$ &256 &512 &640 &1024\\
				Reduction Ratio $R$ &8 &4 &2 &1 \\
				Patch Window Size ($K,S,P$) &(7,4,3) &(3,2,1) &(3,2,1) &(3,2,1) \\
				Number of Transformers $L$  &2 &2 &2 &2   \\		
				\hline
		\end{tabular}}
		\label{tab::rgbformer}
	\end{table}

% Point clouds are irregular, which can not be processed by powerful deep learning models, such as convolutional neural networks directly.Point-based methods such as PointNet and its variants, consume raw points directly to learn 3D representations, which mitigates the drawback of converting point clouds to some regular structures. Nevertheless, due to the irregularity of point cloud data, point-based learning operations have to be permutation-invariant and adaptive to the input size. To achieve this, it learns simple symmetric functions (\textit{e.g}, using pointwise feedforward networks with pooling functions) which highly restricts its representation power. However, the straightforward application of transformer to 3D point clouds is prohibitively expensive because the computation cost grows quadratically with the input size.

%The architecture of Pointformer follow the same design principles as Pixelformer, Due to the irregular properties of point clouds, we utilize MLP to replace the 2D convolutional operation in the feed-forward network.
% to model long-range contextural correlation among points.
\subsection{Instance Representation Learning on Point Clouds}
	\label{sec:pointformer}

Due to the disorder and irregularity of 3D point clouds, deep representation learning on point clouds is very challenging. Moreover, the operations on point clouds must be permutation-invariant, which makes this problem more difficult~\cite{zhang2020pointwise}. The recent transformer architecture~\cite{vaswani2017attention} and its core component, the self-attention mechanism, not only meet the demand for permutation-invariance but also have been proven to be highly expressive in modeling the long-range dependencies between different elements within a sequence of data. Therefore, adapting the transformer architectures to point cloud processing is an ideal choice. Nevertheless, only a few studies~\cite{zhao2021pointtransformer,guo2021pct,pan20213d} have attempted this, and the existing networks suffer from growing memory requirements, as well as high computational consumption. To address this problem, we design a simple but effective transformer-based 3D point cloud learning network called Pointformer suitable for learning the geometric representations of the object instance. %In the experiments, we found that this simple design is sufficient to our object pose estimation task.

As illustrated in \figurename~\ref{fig:fig_pcformer}, given an input point cloud $\mathcal{P}$ with $N$ points, before the main network starts, we first apply two layers of shared MLPs to map the raw points to the $d$-dimensional feature description. Then, similar to Pixelformer, our Pointformer is an encoder-decoder network that contains two main components: 1) a cascaded transformer encoder that includes $I$ ($I$ = 4) stages with different dimensions to capture multilevel point embeddings and 2) an all-MLP decoder to unify and fuse the multilevel embeddings to produce the pointwise geometric representations of the object instance.

\subsubsection{Cascaded Transformer Encoder}

At each stage $i$, where $i\in \{1,2,3,4\}$, there is an input embedding layer (IFE) that maps the $d$-dimensional feature description to the corresponding embedded dimension $C_i$, followed by two layers of alternating channelwise multihead attention (CWMHA) and a FFN to model the long-range context interactions among points and to further improve the point representations.

\paragraph{Input Feature Embedding}

There are many input feature embedding methods in the literature~\cite{guo2021pct}; these are not covered in this work. For the sake of simplicity, we directly apply an MLP layer to the $d$-dimensional feature description to obtain pointwise embeddings with dimensions of $C_i$, where $i\in \{1,2,3,4\}$. Thus, we can obtain multilevel point embeddings with the shape of $\{\mathcal{F}_i|i=1,2,3,4\}$ = \{$N \times C_1$, $N \times C_2$, $N \times C_3$, $N \times C_4$\}.

\paragraph{Channelwise Multihead Attention}
To highlight the interactions across different embedding channels, we build a CWMHA module to enhance the contextual point representations. Specifically, let $Q_i$, $K_i$, $V_i$ = ${\rm MLP}(C_i, C_i)(\mathcal{F}_i)$, $i \in \{1,2,3,4\}$ be the \textit{query}, \textit{key} and \textit{value} vectors at each stage $i$, and each head of $Q_i$, $K_i$, and $V_i$ has the same shape of $N\times d_{head(i)}$, where $d_{head(i)}=C_i / M_i$, $M_i$ denotes the number of attention heads, and $N$ is the number of points. Therefore, the CWMHA operation is formally defined as

\begin{align}
		\label{eq:att_pc} A_i = {\rm{Softmax}}(\frac{Q_iK_i^T}{\sqrt{d_{head(i)}}}), \forall i, \\
		\label{eq:sa_pc} {\rm{Attention}}(Q_i,K_i,V_i)= A_iV_i, \forall i,	
	\end{align}

\noindent where $A_i \in \mathbb{R}^{C_i\times C_i}$ denotes the attention weights between channels, which captures the channelwise importance. Since $Q_i$, $K_i$, and $V_i$ in Eq.~\eqref{eq:att_pc} are determined by linear transformations and the input multilevel point embeddings $\mathcal{F}_i$, $i\in \{1,2,3,4\}$, they are all order-independent. In addition, the softmax function in Eq.~\eqref{eq:att_pc} and the weighed sum operation in Eq.~\eqref{eq:sa_pc} are both permutation-invariant. Therefore, the CWMHA mechanism is permutation-invariant, making it suitable for disordered and irregular point cloud processing. Without abuse of notation, we denote the multilevel features after CWMHA as \{$\mathcal{F}_i|i=1,2,3,4$\}, the same as the input.

% By leveraging this mechanism, we can encode much wider range of channel-wise interactions into the instance geometry representations.
\paragraph{Feed-Forward Network}

We utilize the instance normalization operation, the GELU~\cite{hendrycks2016gaussian} activation function, and MLPs to form the FFN in our Pointformer. The process is defined as

\begin{align}
		\hat{\mathcal{F}}^0_i &= {\rm MLP}(C_i, C_{expand})({\rm IN}(\mathcal{F}_i)), \forall i, \\
		\hat{\mathcal{F}}^1_i &= {\rm MLP}(C_{expand}, C_{expand})(\hat{\mathcal{F}}^0_i), \forall i, \\
		\hat{\mathcal{F}}^2_i &= {\rm GELU}(\hat{\mathcal{F}}^1_i), \forall i, \\
		\hat{\mathcal{F}}_i &= {\rm MLP}(C_{expand}, C_i)(\hat{\mathcal{F}}^2_i) + \mathcal{F}_i, \forall i.
	\end{align}

where $C_{expand}$ represents the expanded embedding dimension of the FFN~\cite{vaswani2017attention} and $\{\hat{\mathcal{F}}_i|i= 1,2,3,4\}$ denote the output multilevel point embeddings of the FFN module. To be consistent with Pixelformer, we set the expansion ratio of each stage as $[8, 8, 4, 4]$.

\subsubsection{All-Multilayer Perceptron Decoder}

We design a simple all-MLP decoder to generate the pointwise geometric representations of the object instance from the multilevel point embeddings created by the cascaded transformer encoder. First, the multilevel point embeddings $\{\hat{\mathcal{F}}_i|i=1,2,3,4\}$ are fed into four separate MLP layers to unify the different channel dimensions to $\mathcal{D}$. Second, the unified point embeddings are concatenated to produce new point embeddings with the shape of $N \times 4\mathcal{D}$. Finally, an MLP layer is applied to project the concatenated point embeddings to a $\mathcal{D}$-dimensional feature description. We formulate the process of the decoder as follows:

\begin{align}
		\hat{\mathcal{F}}^0_i &= {\rm MLP}(C_i,\mathcal{D})(\hat{\mathcal{F}}_i), \forall i, \\
		\hat{\mathcal{F}}^1 &= {\rm Concat}(\hat{\mathcal{F}}^0_i), \forall i,\\
		\mathcal{F}       &= {\rm MLP}(4\mathcal{D},\mathcal{D})(\mathcal{\hat{F}}^1),		        
	\end{align}

\noindent where $\mathcal{F}$ refers to the output pointwise geometric embedding with the shape of $N \times \mathcal{D}$ and $\mathcal{D}$ is the output dimension, which is set as 64 to be consistent with previous geometric extractors~\cite{tian2020shape,wang2019densefusion}. The design details of Pointformer can be found in \tablename~\ref{tab::pcformer}.

% p{3.1cm}<{\centering}|p{0.5cm}<{\centering} p{0.5cm}<{\centering}p{0.6cm}<{\centering}p{0.5cm}<{\centering}p{0.6cm}<{\centering}p{0.6cm}<{\centering}p{1.13cm}<{\centering}|p{0.5cm}<{\centering}p{0.5cm}<{\centering}p{0.6cm}<{\centering}
\begin{table}[!ht]
		\centering
		\renewcommand{\multirowsetup}{\centering} 
		\caption{Detailed settings of the proposed Pointformer for instance representation learning on point clouds.} 
		\resizebox{0.5\textwidth}{!}{\begin{tabular}{l|p{0.7cm}<{\centering}p{0.7cm}<{\centering}p{0.7cm}<{\centering}p{0.7cm}<{\centering}}
				\hline
				\multirow{2}{2cm}{Param Name} &\multicolumn{4}{c}{Stage $I$}  \\\cline{2-5} 
				&1 &2 &3 &4 \\
				\hline
				Number of Attention Heads $M$ &1 &2 &5 &8\\
				FFN Expansion Ratio $R_{FFN}$ &8 &8 &4 &4\\
				Feature Channels $C$ &32 &64 &160 &256\\
				Expanded Dimension $C_{expand}$ &256 &512 &640 &1024\\
				Number of Transformers $L$ &2 &2 &2 &2   \\	
				\hline
		\end{tabular}}
		\label{tab::pcformer}
	\end{table}

\subsection{Joint and Dense Representation Generation and Pose Estimation}
	\label{sec:dense_fusion}

\begin{figure}[!ht]
		\centering
		\includegraphics[scale=0.75]{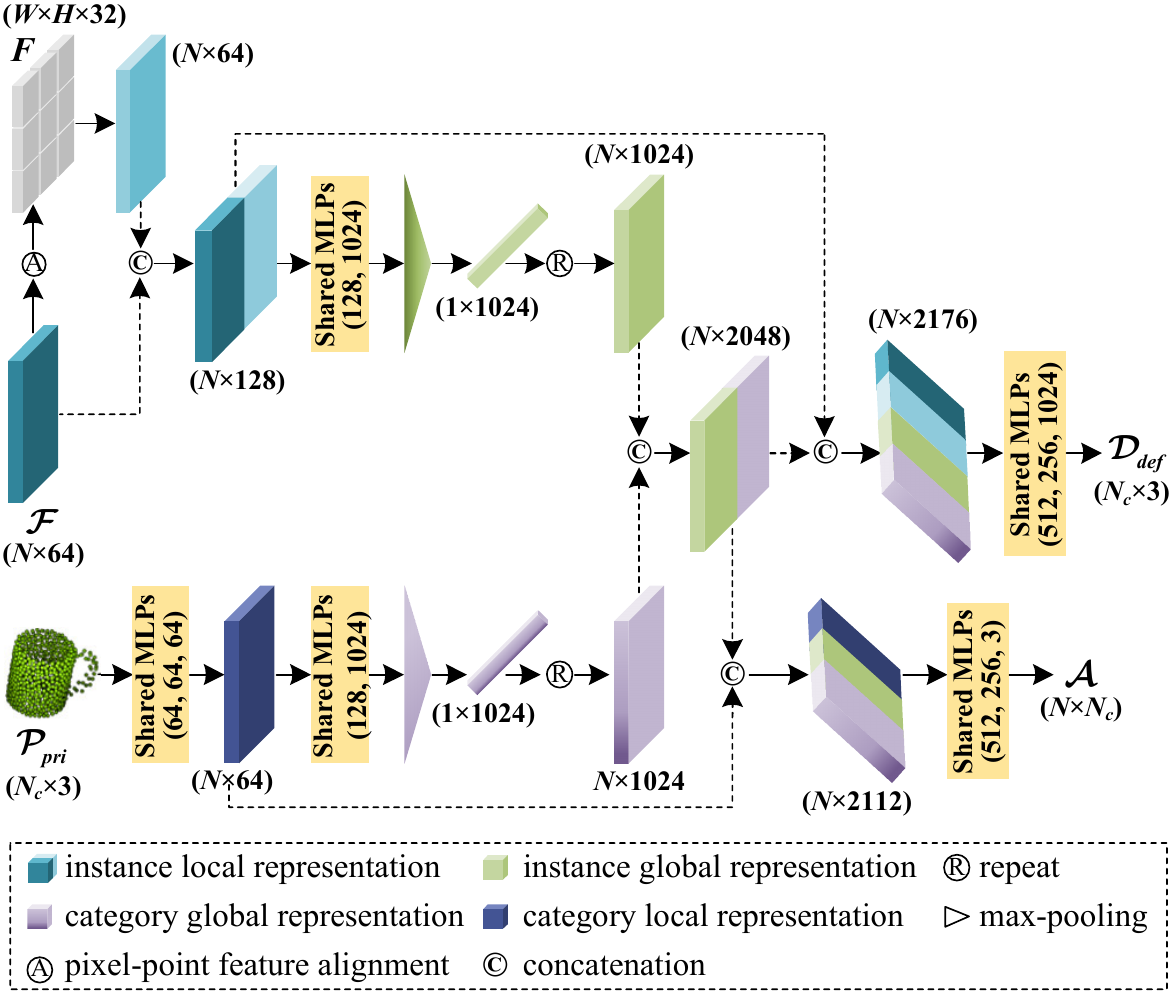}
\caption{The structure of the MSA network. The network takes the pixelwise appearance representations, the pointwise geometric representations, and the categorical shape priors as input and outputs the dense instance representations (\textit{e.g}., correspondence matrix and deformation field) used to calculate the 6D object pose.}
		\label{fig:fig_cma}
\end{figure}

We utilize an MSA network to unify and to generate the dense instance representations, as suggested in SPD~\cite{tian2020shape} (\textit{i.e}., deformation field and correspondence matrix) regarding the pixelwise appearance representations $F\in \mathbb{R}^{H\times W\times D}$, the pointwise geometric representations $\mathcal{F}\in \mathbb{R}^{N\times \mathcal{D}}$, and the categorical shape priors $\mathcal{P}_{pri} \in \mathbb{R}^{N_c\times 3}$ provided by SPD~\cite{tian2020shape}, where $N_c$ is the number of points in the shape priors.

As shown in \figurename~\ref{fig:fig_cma}, the MSA contains two parallel branches. In the upper branch, we associate the appearance representations $F$ with the geometric representations $\mathcal{F}$ according to the natural pixel-to-point correspondence. In addition, to unify the feature dimensions of these two modalities, an MLP layer is applied. After that, the aligned cross-modality representations are concatenated and termed \textit{instance local representation} and fed into shared MLPs, followed by an average pooling layer to obtain the \textit{instance global representation}. In the lower branch, the categorical shape priors $\mathcal{P}_{pri}$ are utilized to provide the prior knowledge of each category. Specifically, we employ two shared MLPs to extract the \textit{category local representation} and the \textit{category global representation} from the shape priors. Then, the two category-level representations are concatenated and enriched either by the \textit{category local representation} followed by shared MLPs to generate the deformation field $\mathcal{D}_{def}\in \mathbb{R}^{N_c\times 3}$ from the categorical shape priors to a particular instance canonical model or by the \textit{instance local representation} followed by other shared MLPs to generate the correspondence matrix $\mathcal{A}\in \mathbb{R}^{N\times N_c}$, which represents the soft correspondence between each point in the input instance point cloud $\mathcal{P}$ and all points in the reconstructed instance model $\hat{\mathcal{P}}$, in which $\hat{\mathcal{P}} = \mathcal{P}_{pri}+\mathcal{D}_{def}$. Finally, the NOCS coordinates of the instance point cloud, denoted as $\hat{\mathcal{M}}$, are obtained by multiplying the correspondence matrix $\mathcal{A}$ and the reconstructed object model $\hat{\mathcal{P}}$, \textit{i.e}.,

\begin{equation}
		%$\hat{\mathcal{P}} &= \mathcal{P}_{pri}+\mathcal{D}_{def}, \\
		\hat{\mathcal{M}} = \mathcal{A}\times \hat{\mathcal{P}}.
	\end{equation}

Given the instance point cloud $\mathcal{P}$ with its corresponding NOCS coordinates $\hat{\mathcal{M}}$, we use the Umeyama algorithm~\cite{umeyama1991least} to estimate the optimal similarity transformation parameters (\textit{a.k.a}., rotation, translation, and scale), where the rotation and translation parameters correspond to the 6D object pose and the scale parameter corresponds to the object size. The RANdom SAmple Consensus (RANSAC) algorithm~\cite{fischler1981random} is also used to remove outliers and to achieve robust estimation.

\subsection{Loss Functions}
	\label{sec:loss}

We utilize the same loss functions as those suggested in SPD~\cite{tian2020shape} to supervise different components of our method.

\subsubsection{Reconstruction Loss}

We employ the Chamfer distance to measure the similarity between the ground truth 3D model and the reconstructed 3D model, which is defined as
\begin{equation}
		\label{eq:cd}
		L_{cd}(\hat{\mathcal{P}}^i_c, \mathcal{P}^i_c)=\sum_{x\in \mathcal{P}^i_c}\mathop{\min}_{y\in \hat{P}^i_c}|| x-y ||^2_2+\sum_{y\in \hat{\mathcal{P}}^i_c}\mathop{\min}_{x\in P^i_c}|| x-y ||^2_2,
	\end{equation}

\noindent where $\mathcal{P}^i_c$ is the ground truth 3D point cloud model of instance $i$ from category $c$ and $\hat{\mathcal{P}}^i_c$ is its corresponding reconstructed 3D point cloud model. Since the reconstructed model is derived by the categorical shape priors and deformation field, \textit{i.e}., $\hat{\mathcal{P}} = \mathcal{P}_{pri} + \mathcal{D}_{def}$, the regression of the deformation field is also implicitly supervised by the reconstruction loss.

\subsubsection{Correspondence Loss}

We follow SPD~\cite{tian2020shape} to indirectly supervise the deformation field $\mathcal{D}_{def}$ by supervising the NOCS coordinates $\hat{\mathcal{M}}$ with the smooth $L_1$ loss function,

\begin{align}
		\hat{{L}}_{cor}(m, m_{gt})= \left\{\begin{matrix}
			5(m-m_{gt})^2,	& |m-m_{gt}|\leq 0.01\\ 
			|m-m_{gt}|-0.05,	& {\rm otherwise}
		\end{matrix}\right.,
	\end{align}

\begin{equation}
		L_{cor}(\hat{\mathcal{M}}, \mathcal{M}) = \frac{1}{N}\sum_{m\in \hat{\mathcal{M}}}\hat{{L}}_{cor}(m, m_{gt}),
	\end{equation}

\noindent where $m_{gt}$ is the ground truth NOCS coordinate from $\mathcal{M}$ and $m$ is the predicted NOCS coordinate from $\hat{\mathcal{M}}$. Therefore, the correspondence loss is computed as the average correspondence loss over all the NOCS coordinates.

\subsubsection{Regularization Losses}

To penalize large deformations and to constrain the correspondence matrix $\mathcal{A}$ to be sparse, we impose two regularization losses on each point $d_i$ in the deformation field and each raw $\mathcal{A}_i$ of the correspondence matrix.

\begin{equation}
		L_{def} = \frac{1}{N_c}\sum_{d_i\in \mathcal{D}_{def}}||d_i||_2,
	\end{equation}

\begin{equation}
		L_{spar} = \frac{1}{N}\sum_i\sum_j-\mathcal{A}_{i,j}\log\mathcal{A}_{i,j}.
	\end{equation}

In summary, the overall loss function is a weighted sum of all four loss functions:

\begin{equation}
		\label{eq:loss}
		L = \lambda_1L_{cd}+\lambda_2L_{cor}+\lambda_3L_{def}+\lambda_4L_{spar}.
	\end{equation}

\section{Experiments}
	\label{sec:experiments}

In this section, we conduct extensive experiments on two state-of-the-art benchmark datasets (\textit{i.e}., CAMERA25~\cite{wang2019normalized} and REAL275~\cite{wang2019normalized}) to evaluate the performance of the presented method and to compare our results with those of the baseline approach~\cite{tian2020shape} and other state-of-the-art methods. We also perform a comprehensive ablation analysis (18 experiments in total) to verify the prospective advantages of the proposed Pixelformer and Pointformer of our framework and to compare each of them with the latest state-of-the-art transformer-based appearance~\cite{wang2021pyramid} or geometric~\cite{zhao2021pointtransformer} representation generators. Furthermore, we show the visualization results of the pose estimation results and the 3D object models reconstructed by our network, which qualitatively demonstrate the effectiveness of our approach.

\begin{table*}
		\renewcommand{\multirowsetup}{\centering} 
		\centering
		\caption{Category-wise results of our network on the CAMERA25 and REAL275 datasets under different evaluation metrics.} 
		
		\begin{tabular}{c|c|ccccccc}
			\hline
			Dataset &Category& \textit{IoU}$_{50}$ & \textit{IoU}$_{75}$ &\ang{5} 2cm& \ang{5} 5cm& \ang{10} 2cm & \ang{10} 5cm &\ang{10} 10cm\\
			\hline
			\multirow{7}{1.5cm}{CAMERA25} &Bottle &0.9242 &0.8524 &0.7357 &0.8962 &0.7591 &0.9436 &0.9760\\
			&Bowl &0.9668 &0.9563 &0.9396 &0.9430 &0.9727 &0.9801 &0.9810\\
			&Camera &0.9114 &0.8228 &0.4942 &0.5018 &0.7109 &0.7374 &0.7383\\
			&Can &0.9181 &0.9108 &0.9508 &0.9552 &0.9632 &0.9741 &0.9760\\
 			&Laptop &0.9576 &0.8619 &0.7131 &0.7801 &0.7790 &0.8897 &0.9316\\
			&Mug &0.9297 &0.9048 &0.5252 &0.5258 &0.7527 &0.7538 &0.7539\\
			&Average &0.9346 &0.8849 &0.7265 &0.7670 &0.8229 &0.8798 &0.8928\\

			\hline
			
			Dataset &Category& \textit{IoU}$_{50}$ &\textit{IoU}$_{75}$ &\ang{5} 2cm& \ang{5} 5cm& \ang{10} 2cm & \ang{10} 5cm &\ang{10} 10cm\\
			\hline
			\multirow{7}{1cm}{REAL275} &Bottle &0.5766 &0.5005 &0.5799 &0.6318 &0.7969 &0.8703 &0.9452\\
			&Bowl &0.9999 &0.9992 &0.7874 &0.8186 &0.9548 &0.9914 &0.9914\\
			&Camera &0.8709 &0.1917 &0.0000 &0.0000 &0.0014 &0.0019 &0.0019\\
			&Can &0.7146 &0.6996 &0.5350 &0.5624 &0.8573 &0.9551 &0.9555\\
			&Laptop &0.8334 &0.6170 &0.3383 &0.4461 &0.6163 &0.9217 &0.9361\\
			&Mug &0.9878 &0.8577 &0.0490 &0.0524 &0.3166 &0.3333 &0.3333\\
			&Average &0.8306 &0.6443 &0.3816 &0.4186 &0.5906 &0.6789 &0.6989\\
			\hline
		\end{tabular}
		\label{tab::category-specific}
	\end{table*}

\begin{figure*}[!ht]
		\centering
		\includegraphics[scale=0.6]{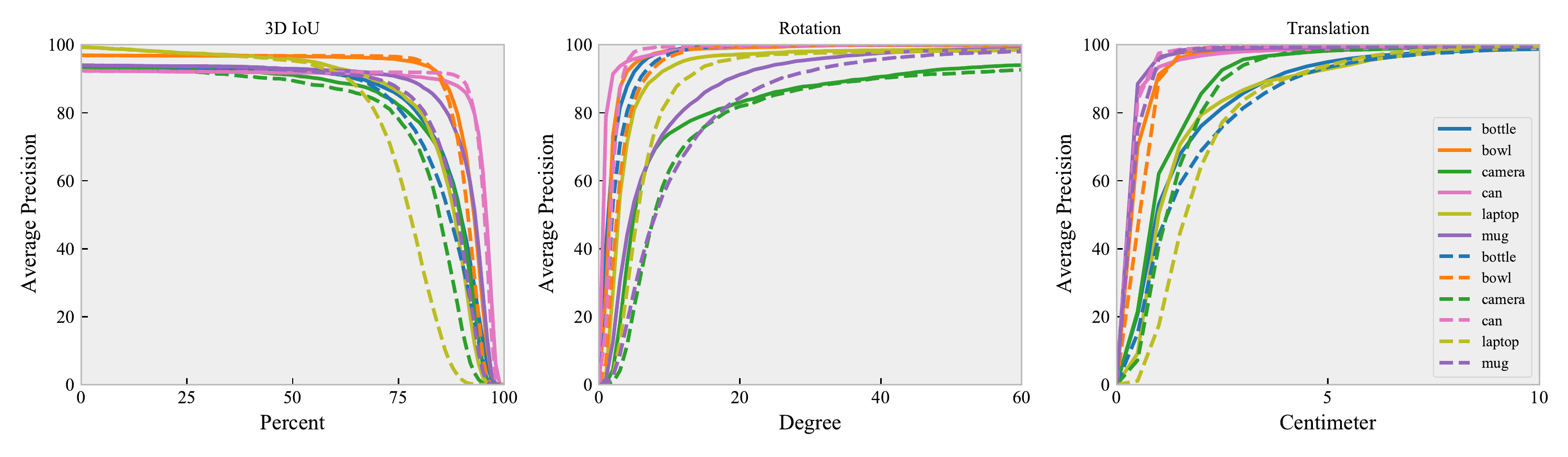}
\caption{Category-level comparison with the baseline method: SPD~\cite{tian2020shape} on the CAMERA25 dataset. The solid lines denote the results of our 6D-ViT, and the dashed lines denote the results of SPD~\cite{tian2020shape}. We report the precision of different thresholds with the 3D IoU, rotation and translation errors.}
		\label{fig:category_camera}
\end{figure*}

\begin{figure*}[!ht]
		\centering
		\includegraphics[scale=0.6]{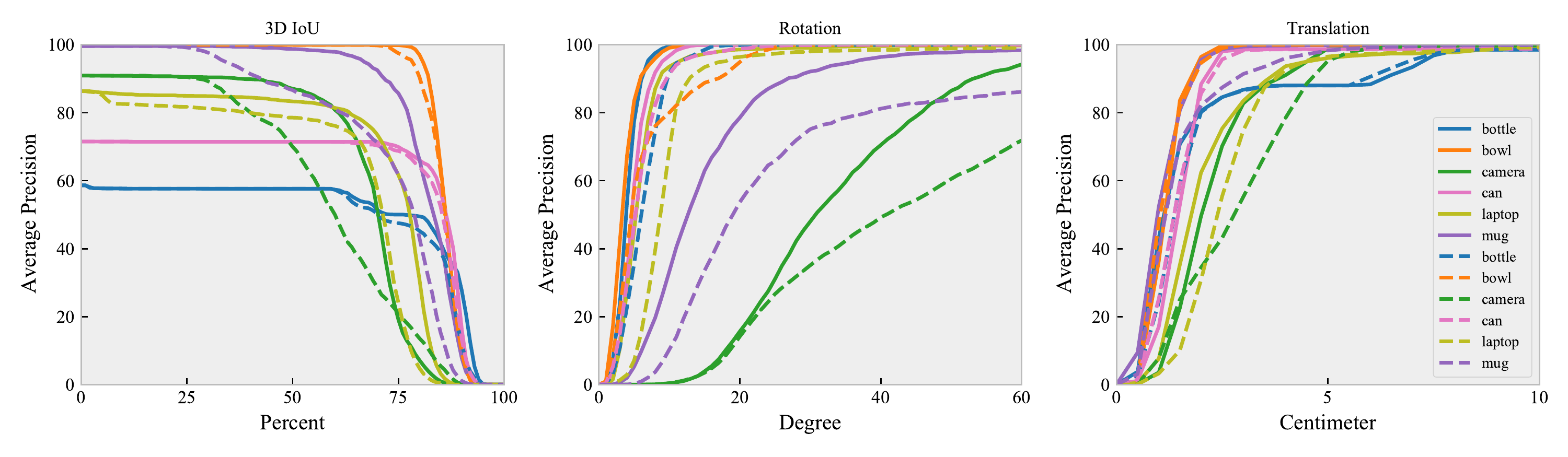}
\caption{Category-level comparison with SPD~\cite{tian2020shape} on the REAL275 dataset. The solid lines denote the results of our 6D-ViT, and the dashed lines denote the results of SPD~\cite{tian2020shape}. We report the precision of different thresholds with the 3D IoU, rotation and translation errors.}
		\label{fig:category_real}
\end{figure*}

\subsection{Datasets}

The CAMERA25 and REAL275 datasets were both compiled by the creators of NOCS~\cite{wang2019normalized}, %Editor: Please ensure that the intended meaning has been maintained in this edit.
with six different categories: bottle, bowl, camera, can, laptop and mug. Specifically, the CAMERA25 dataset was generated by rendering and compositing synthetic object instances from ShapeNet-Core~\cite{chang2015shapenet} under different views. The training set covers 1,085 object instances, while the evaluation set covers 184 different instances. In total, it contains 300K composite RGB-D images, where 25K are set aside for evaluation. The REAL275~\cite{wang2019normalized} dataset is complementary to the CAMERA25 dataset. It includes 4,300 real-world images of 7 scenes for training and 2,750 real-world images of 6 scenes with 3 unseen instances per category for evaluation. Both the training set and testing set contain 18 real object instances spanning the 6 categories.

\subsection{Metrics}

Following~\cite{wang2019normalized,tian2020shape,chen2020category,chen2020learning,chen2021fs,lin2021dualposenet}, we independently evaluate the performance of our method in terms of 3D object detection and 6D object pose estimation and compare the results with those of the state-of-the-art methods. For 3D object detection, we report the average precision at different intersection over union thresholds 3D$_X$. For 6D object pose estimation, the average precision is computed as $n\degree m$cm. We ignore the rotational error around the axis of symmetry for symmetrical object categories (\textit{e.g}. bottle, bowl, and can). In particular, we regard mugs as symmetrical objects when the handle is absent and asymmetric objects otherwise. In addition, the Chamfer distance (Eq.~\eqref{eq:cd}) is employed to evaluate the accuracy of shape reconstruction from single-view RGB-D images.

% \resizebox{1\textwidth}{!}{
\begin{table*}[!ht]
		\renewcommand{\multirowsetup}{\centering} 
		\centering
		\caption{Quantitative comparisons on the CAMERA25 and REAL275 datasets. We report the mAP \textit{w.r.t}. different thresholds on the 3D IoU and rotation and translation errors. The results of the comparison methods are summarized from their original papers.} 
		
		\resizebox{1\textwidth}{!}{\begin{tabular}{p{3.38cm}<{\centering}|p{0.45cm}<{\centering} p{0.45cm}<{\centering}p{0.5cm}<{\centering}p{0.5cm}<{\centering}p{0.6cm}<{\centering}p{0.6cm}<{\centering}p{1.13cm}<{\centering}|p{0.45cm}<{\centering}p{0.45cm}<{\centering}p{0.5cm}<{\centering}p{0.5cm}<{\centering}p{0.6cm}<{\centering}p{0.6cm}<{\centering}p{1.13cm}<{\centering}}
				\hline
				\multirow{2}{1cm}{Method}  &\multicolumn{7}{c}{CAMERA25}  &\multicolumn{7}{c}{REAL275} \\\cline{2-8} \cline{9-15}
				&3D$_{50}$ & 3D$_{75}$ &\ang{5}2cm& \ang{5}5cm& \ang{10}2cm & \ang{10}5cm &\ang{10}10cm &3D$_{50}$ &3D$_{75}$ &\ang{5}2cm& \ang{5}5cm& \ang{10}2cm & \ang{10}5cm &\ang{10}10cm\\

				\hline
				NOCS [CVPR'19]~\cite{wang2019normalized}  & 83.9\%& 69.5\% &32.3\% &40.9\% &48.2\% &64.6\% &- &78.0\% &30.1\% &7.2\% &10.0\% &13.8\% &25.2\% &26.7\%\\
				CASS [CVPR'20]~\cite{chen2020learning}  &-&-&-&-&-&-&- &77.7\% &- &- &23.8\% &- &58.0\% &58.3\%\\
				NOF [ECCV'20]~\cite{chen2020category}  &-&-&-&-&-&-&- &76.9\% &30.1\% &7.2\% &9.8\% &13.8\% &24.0\% &24.3\%\\
				
				SPD [ECCV'20]~\cite{tian2020shape} &93.2\% &83.1\% &54.3\% &59.0\% &73.3\% &81.5\% &- &77.3\% &53.2\% &19.3\% &21.4\% &43.2\% &54.1\% &-\\
				%COM[SIVP'21]~\cite{sun2021efficient} &85.4\% &- &- &42.2\% &- &65.8\% &66.7\% &78.3\% &- &- &11.0\% &- &26.4\% &27.1\%\\
				FS-Net [CVPR'21]~\cite{chen2021fs} &-&-&-&-&-&-&- &\textbf{92.2\%} &63.5\% &- &28.2\% &- &60.8\% &64.6\%\\
				DualPoseNet [ICCV'21]~\cite{lin2021dualposenet}  &92.4\% &86.4\% &64.7\% &70.7\% &77.2\% &84.7\% &- &79.8\%  &62.2\% &29.3\% &35.9\% &50.0\% &66.8\% &-\\
				
				6D-ViT [Ours] &\textbf{93.5\%} &\textbf{88.5\%} &\textbf{72.6\%} &\textbf{76.7\%} &\textbf{82.3\%} &\textbf{88.0\%} &\textbf{89.3\%} &83.1\% &\textbf{64.4}\% &\textbf{38.2}\% &\textbf{41.9\%} &\textbf{59.1\%} &\textbf{67.9\%} &\textbf{69.9\%}\\		
				\hline
		\end{tabular}}
		\label{tab::quantitative}
	\end{table*}

%fig7_mAP_mean_ablation_camera
\begin{figure*}[!ht]
		\centering
		\includegraphics[scale=0.6]{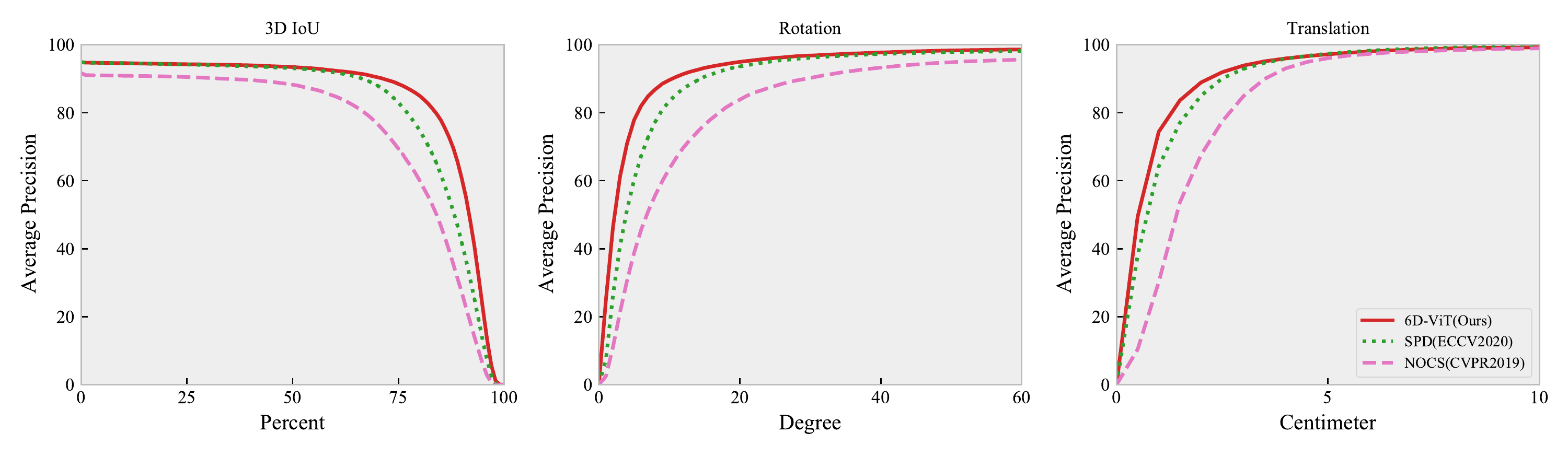}
\caption{The average precision of all categories on the CAMERA25 dataset. We present different thresholds with the 3D IoU, rotation and translation errors.}
		\label{fig:mean_comparison_camera}
\end{figure*}

\begin{figure*}[!ht]
		\centering
		\includegraphics[scale=0.8]{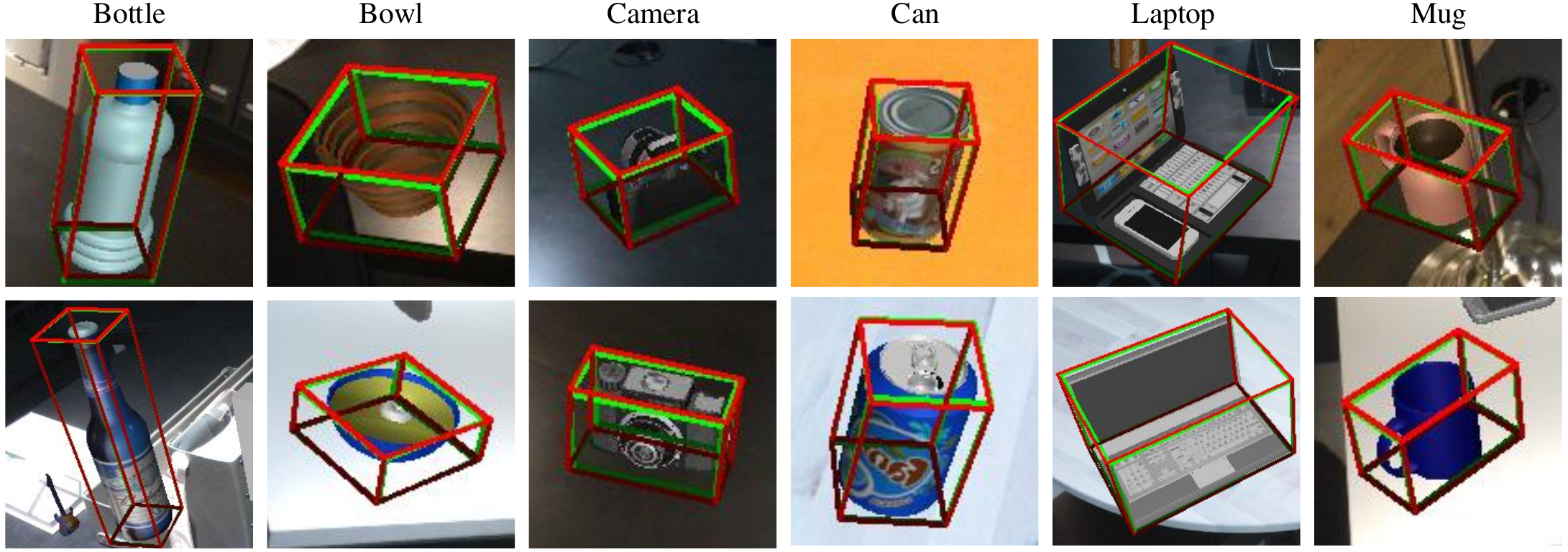}
\caption{Visualization of 6D object poses on the CAMERA25 dataset. Green bounding boxes denote the ground truth. Red boxes denote our estimations. Our results match the ground truth well in terms of both pose and size.}
		\label{fig:visualization_camera}
\end{figure*}

\subsection{Implementation Details}
We use the PyTorch~\cite{paszke2019pytorch} framework to implement our method, and the state-of-the-art framework SPD~\cite{tian2020shape} is developed as the baseline. All the experiments are conducted on a PC with an i9-10900K 3.70 GHz CPU and two GTX 2080Ti GPUs. The RGB images are cropped and resized to 256$\times$256. The number of points in the instance point clouds and categorical shape priors is set as 1,024. We use Adam~\cite{kingma2014adam} to optimize our network, where the initial learning rate is set as 1e-4, and we halve it every 20 epochs with a weight decay of 1e-6. The maximum number of epochs is 100. For the hyperparameters of the loss function, we follow SPD~\cite{tian2020shape} and set $\lambda_1=5.0$, $\lambda_2=1.0$, $\lambda_3=1.0$, and $\lambda_4=1e-4$.

% (Eq.~\eqref{eq:loss})

\subsection{Evaluation of the 6D Object Pose Estimation}
	\label{sec:eva_pose}

        We evaluate the performance of our 6D object pose estimation network, 6D-ViT, on two widely used benchmark datasets (\textit{i.e.}, CAMERA25 and REAL275) and compare our results with those of the baseline method, SPD~\cite{tian2020shape}, and five other state-of-the-art methods, including NOCS~\cite{wang2019normalized}, CASS~\cite{chen2020learning}, NOF~\cite{chen2020category}, % Please note that the algorithm in this reference is never referred to as "NOF" in the original paper.
FS-Net~\cite{chen2021fs}, and DualPoseNet~\cite{lin2021dualposenet}. %As far as we know, these are all the popular methods in the current literature to solve the challenging category-level 6D object pose estimation problem. %We report the quantitative and qualitative results in Section~\ref{sec:quantitative} and Section~\ref{sec:qualitative}, respectively.

\begin{figure*}[!ht]
		\centering
		\includegraphics[scale=0.6]{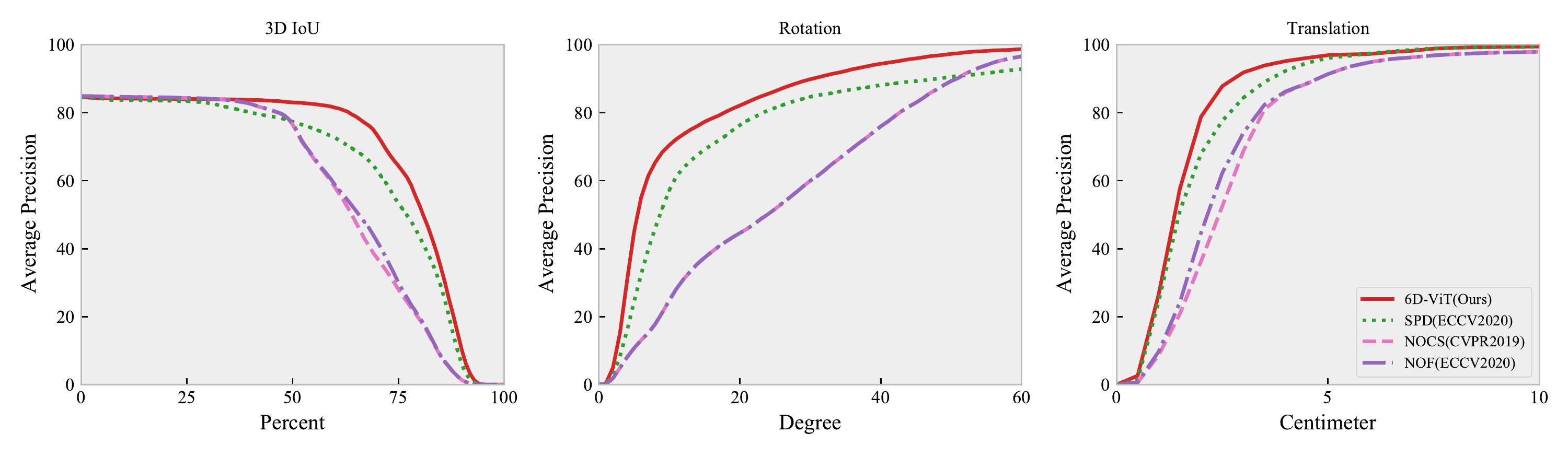}
\caption{The average precision of all categories for the REAL275 dataset. We present different thresholds with the 3D IoU, rotation and translation errors. }
		\label{fig:mean_comparison_real}
\end{figure*}

\begin{figure*}[!ht]
		\centering
		\includegraphics[scale=0.63]{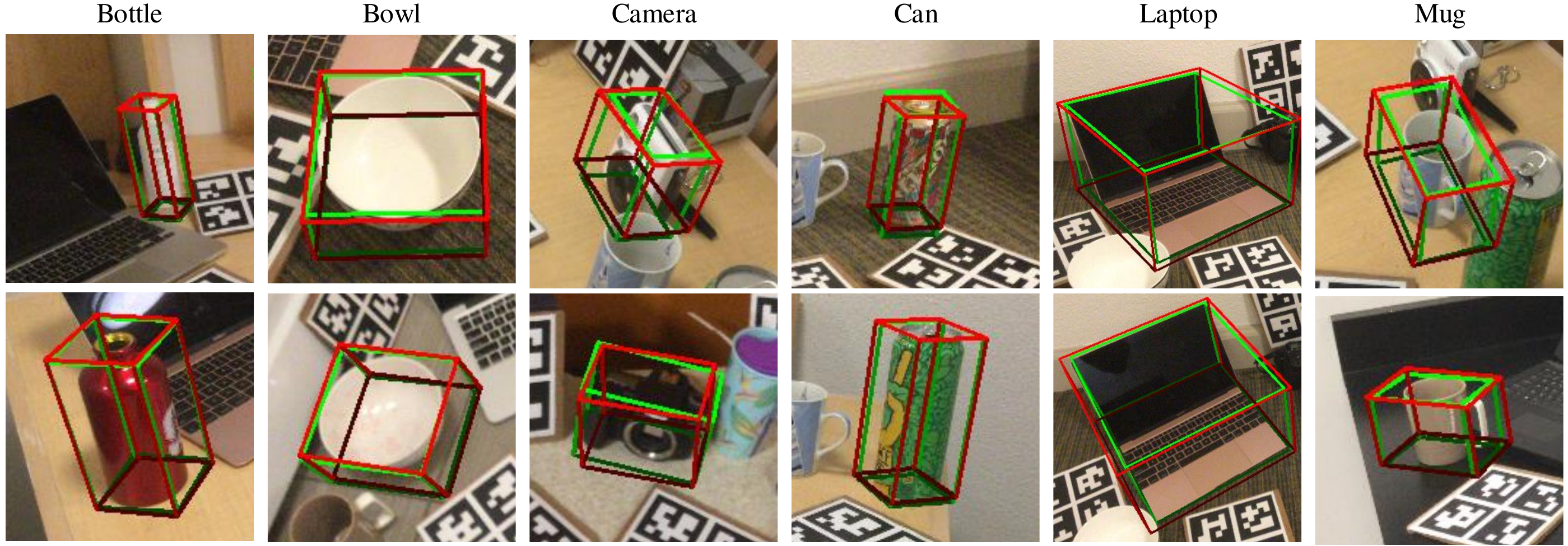}
\caption{Visualization of the 6D object poses for the REAL275 dataset. Green bounding boxes denote the ground truth. Red boxes denote our estimations. Our results match the ground truth well in terms of both pose and size.}
		\label{fig:visualization_real}
\end{figure*}

\subsubsection{Comparison with the Baseline}

To verify the effectiveness of our method, we report the category-specific precision of our 6D-ViT on the CAMERA25 and REAL275 datasets under different thresholds with the 3D IoU and rotation and translation errors and compare all results with those of SPD~\cite{tian2020shape}. The comparisons can be found in \figurename~\ref{fig:category_camera} and \figurename~\ref{fig:category_real}, respectively. The results show that for all evaluation metrics, our method achieves remarkably better performance than the baseline method in each category of the two datasets. In addition, we provide detailed category-specific results under several widely used thresholds to more intuitively show our results. The results are provided in \tablename~\ref{tab::category-specific}.

\begin{table*}
		\renewcommand{\multirowsetup}{\centering} 
		\centering
		\caption{Evaluation of the 3D model reconstruction using the CD metric ($\times$10$^{-3}$).} 
		\resizebox{1\textwidth}{!}{\begin{tabular}{p{1.8cm}<{\centering}|p{0.7cm}<{\centering}p{0.7cm}<{\centering}p{0.7cm}<{\centering}p{0.65cm}<{\centering}p{0.7cm}<{\centering}p{0.7cm}<{\centering}p{0.8cm}<{\centering}|p{0.7cm}<{\centering}p{0.7cm}<{\centering}p{0.7cm}<{\centering}p{0.65cm}<{\centering}p{0.7cm}<{\centering}p{0.7cm}<{\centering}p{0.8cm}<{\centering}}
				\hline
				\multirow{2}{1.8cm}{Method}  &\multicolumn{7}{c}{CAMERA} &\multicolumn{7}{c}{REAL275} \\\cline{2-8} \cline{9-15}
				&Bottle &Bowl &Camera &Can &Laptop &Mug &Average &Bottle &Bowl &Camera &Can &Laptop &Mug &Average\\
				\hline
				Baseline~\cite{tian2020shape} &1.81 &1.63 &4.02 &0.97 &1.98 &1.42 &1.97 &3.44 &1.21 &8.89 &\textbf{1.56} &2.91 &\textbf{1.02} &3.17\\
				
				Ours &\textbf{1.38}&\textbf{1.18}	&\textbf{2.86}	&\textbf{0.90}	&\textbf{1.24}	&\textbf{1.17}	&\textbf{1.46} &\textbf{2.37} &\textbf{1.14} &\textbf{6.06} &1.60 &\textbf{1.39} &1.06 &\textbf{2.27}\\
				
				\hline
		\end{tabular}}
		\label{tab::shape}
	\end{table*}

\subsubsection{Comparison with the State-of-the-Art Algorithms}

To further validate the competitiveness of our method, we compare the average precision of 6D-ViT with that of NOCS~\cite{wang2019normalized}, CASS~\cite{chen2020learning}, NOF~\cite{chen2020category}, SPD~\cite{tian2020shape}, FS-Net~\cite{chen2021fs}, and DualPoseNet~\cite{lin2021dualposenet} on both the CAMERA25 and REAL275 datasets. The results are reported in \tablename~\ref{tab::quantitative}.

\paragraph{CAMERA25}

As shown on the left of \tablename~\ref{tab::quantitative}, our 6D-ViT is superior to all existing methods on the CAMERA25 dataset. Specifically, for the 3D object detection metrics 3D$_{50}$ and 3D$_{75}$, our 6D-ViT outperforms the previous best method, DualPoseNet~\cite{lin2021dualposenet}, by 1.1\% and 2.1\%, respectively. In terms of the 6D pose metrics $\ang{5}$2cm, $\ang{5}$5cm, $\ang{10}$2cm and $\ang{10}$10cm, 6D-ViT has advantages of 7.9\%, 6.0\%, 5.1\%, and 3.3\%, respectively, performing significantly better than DualPoseNet~\cite{lin2021dualposenet}. Furthermore, 6D-ViT achieves an average precision of 89.3\% under the $\ang{10}$10cm metric. We show more detailed quantitative comparisons in \figurename~\ref{fig:mean_comparison_camera}, and our qualitative results on the CAMERA25 dataset are shown in \figurename~\ref{fig:visualization_camera}.

\paragraph{REAL275}

As shown on the right of \tablename~\ref{tab::quantitative}, our proposed 6D-ViT is superior to the existing methods under all evaluation metrics except for 3D$_{50}$, which is a rather rough metric. Specifically, for 3D object detection, FS-Net~\cite{chen2021fs} is the current best method in terms of the 3D$_{50}$ metric, and our 6D-ViT achieves the second best average precision, 83.1\%, which is 9.1\% lower than that of FS-Net~\cite{chen2021fs}. However, for the more difficult 3D$_{75}$ metric, the performance of our 6D-ViT is 0.9\% higher than that of FS-Net~\cite{chen2021fs}. For the 6D pose evaluation metrics $\ang{5}$2cm, $\ang{5}$5cm, $\ang{10}$2cm and $\ang{10}$10cm, our 6D-ViT significantly outperforms the previous best method, DualPoseNet~\cite{lin2021dualposenet}, by 8.9\%, 6.0\%, 9.1\%, and 1.1\%, respectively. In addition, 6D-ViT achieves an average accuracy of 69.9\% under the $\ang{10}$10cm metric, which is 5.3\% higher than the best result previously reported.

\begin{table}
		\centering
		\renewcommand{\multirowsetup}{\centering} 
		\caption{The alternative instance appearance feature extractors and instance geometric feature extractors used to verify the efficacy of individual components of our network.} 
		\begin{tabular}{l|p{1.8cm}<{\centering}p{1.8cm}<{\centering}p{1.8cm}<{\centering}}
			\hline
			Appearance &PSP~\cite{tian2020shape}  &PVT~\cite{wang2021pyramid} &PIF (Ours)\\
			Geometry &MLPs~\cite{tian2020shape}  &PT~\cite{zhao2021pointtransformer} &POF (Ours)\\
			
			\hline
		\end{tabular}
		\label{tab::ablation_pair}
	\end{table}

\begin{table*}
		\renewcommand{\multirowsetup}{\centering} 
		\centering
		\caption{Ablation studies on the CAMERA25 and REAL275 datasets. ``PSP", ``MLP",  ``PVT", ``PT", ``PIF", and ``POF" refer to ``PSPNet in the baseline~\cite{tian2020shape}", ``linear projection layers in the baseline~\cite{tian2020shape}", ``pyramid vision transformer~\cite{wang2021pyramid}", ``point transformer~\cite{zhao2021pointtransformer}", ``Pixelformer in this work", and ``Pointformer in this work", respectively.} 
		\resizebox{1\textwidth}{!}{\begin{tabular}{p{0.05cm}<{\centering} p{2.82cm}<{\centering}|p{0.4cm}<{\centering}p{0.4cm}<{\centering}p{0.6cm}<{\centering}p{0.5cm}<{\centering}p{0.6cm}<{\centering}p{0.6cm}<{\centering}p{1.13cm}<{\centering}|p{0.4cm}<{\centering}p{0.4cm}<{\centering}p{0.6cm}<{\centering}p{0.5cm}<{\centering}p{0.6cm}<{\centering}p{0.6cm}<{\centering}p{1.13cm}<{\centering}}
				\hline
				& \multirow{2}{2.8cm}{Method} &\multicolumn{7}{c}{CAMERA25}  &\multicolumn{7}{c}{REAL275} \\\cline{3-9} \cline{10-16}
				& &3D$_{50}$ & 3D$_{75}$ &\ang{5}2cm& \ang{5}5cm& \ang{10}2cm & \ang{10}5cm &\ang{10}10cm &3D$_{50}$ &3D$_{75}$ &\ang{5}2cm& \ang{5}5cm& \ang{10}2cm & \ang{10}5cm &\ang{10}10cm \\
				\hline
				\textcircled{1} & PSP~\cite{tian2020shape} / MLPs~\cite{tian2020shape} &93.2\% &83.1\% &54.3\% &59.0\% &73.3\% &81.5\%  &- &77.3\% &53.2\% &19.3\% &21.4\% &43.2\% &54.1\%  &-\\
				\textcircled{2}  &PVT~\cite{wang2021pyramid} / MLPs~\cite{tian2020shape} &91.2\% &77.9\% &43.3\% &47.0\% &62.7\% &71.2\% &72.3\% &69.6\% &41.8\% &15.4\% &16.5\% &39.5\% &44.1\% &46.0\% \\
				\textcircled{3}  &PVT~\cite{wang2021pyramid} / PT~\cite{zhao2021pointtransformer} &91.4\% &84.2\% &60.0\% &64.2\% &75.5\% &82.2\% &83.2\% &80.2\% &56.8\% &27.8\% &31.4\% &48.8\% &60.5\% &62.3\%\\
				\textcircled{4} &PVT~\cite{wang2021pyramid} / POF (Ours) &91.1\% &83.9\% &63.6\% &67.6\% &77.0\% &82.7\% &84.1\% &76.2\% &53.4\% &32.1\% &35.5\% &51.7\% &61.3\% &63.3\%\\
				\textcircled{5} &PSP~\cite{tian2020shape} / PT~\cite{zhao2021pointtransformer} &93.3\% &87.6\% &60.9\% &65.6\% &77.9\% &85.4\% &86.6\% &79.5\% &58.9\% &29.0\% &32.1\% &52.8\% &61.1\% &62.9\% \\
				\textcircled{6} &PIF (Ours) / PT~\cite{zhao2021pointtransformer} &92.6\% &85.5\% &65.3\% &69.4\% &78.2\% &84.9\% &86.4\%  &81.4\% &\textbf{67.0\%} &35.7\% &38.9\% &58.9\% &67.7\% &69.6\% \\
				
				\textcircled{7}  &PSP~\cite{tian2020shape} / POF (Ours)  &93.4\% &88.0\% &69.5\% &73.8\% &81.1\% &87.1\% &88.6\% &78.4\% &61.4\% &36.1\% &39.1\% &55.9\% &65.1\% &67.1\%\\
				
				\textcircled{8}  &PIF (Ours) / MLPs~\cite{tian2020shape}  &\textbf{93.6\%} &87.9\% &63.6\% &67.8\% &79.7\% &86.5\% &87.5\% &82.4\% &61.6\% &27.8\% &30.1\% &52.4\% &63.6\% &65.6\% \\
				\textcircled{9} &PIF (Ours) / POF (Ours)  &93.5\% &\textbf{88.5\%} &\textbf{72.6\%} &\textbf{76.7\%} &\textbf{82.3\%} &\textbf{88.0\%} &\textbf{89.3\%}&\textbf{83.1\%} &64.4\% &\textbf{38.2\%} &\textbf{41.9\%} &\textbf{59.1\%} &\textbf{67.9\%} &\textbf{69.9\%}\\	
				\hline
		\end{tabular}}
		\label{tab::ablation}
	\end{table*}

% This phenomenon is consistent with the general conclusion that segmentation is more accurate for object localization than detection.
We believe that the performance gap between our method and FS-Net~\cite{chen2021fs} in 3D object detection occurs because FS-Net~\cite{chen2021fs} utilizes YOLOv3~\cite{farhadi2018yolov3} to crop object instances from RGB-D images, while our method, along with other methods, employs Mask-RCNN~\cite{he2017mask} instead. More specifically, YOLOv3 has a higher success rate for coarse instance detection, while Mask R-CNN yields more accurate instance localization. Therefore, FS-Net~\cite{chen2021fs} performs better than our 6D-ViT in terms of the 3D$_{50}$ metric due to the effectiveness of object detection, while our 6D-ViT outperforms FS-Net~\cite{chen2021fs} in terms of the harder 3D$_{75}$ metric, which validates the advantage of our instance representation learning framework. We show more detailed quantitative comparisons on the REAL275 dataset in \figurename~\ref{fig:mean_comparison_real}, and our qualitative results are provided in \figurename~\ref{fig:visualization_real}.

In summary, our proposed instance representation learning network achieves the best performance in both 6D object pose estimation and 3D object detection of the existing category-level pose estimation networks. This excellent performance further reflects the high competitiveness of our 6D-ViT.

\subsection{Evaluation of the Model Reconstruction}
	\label{sec:eva_recons}

\begin{figure*}[!ht]
		\centering
		\includegraphics[scale=0.85]{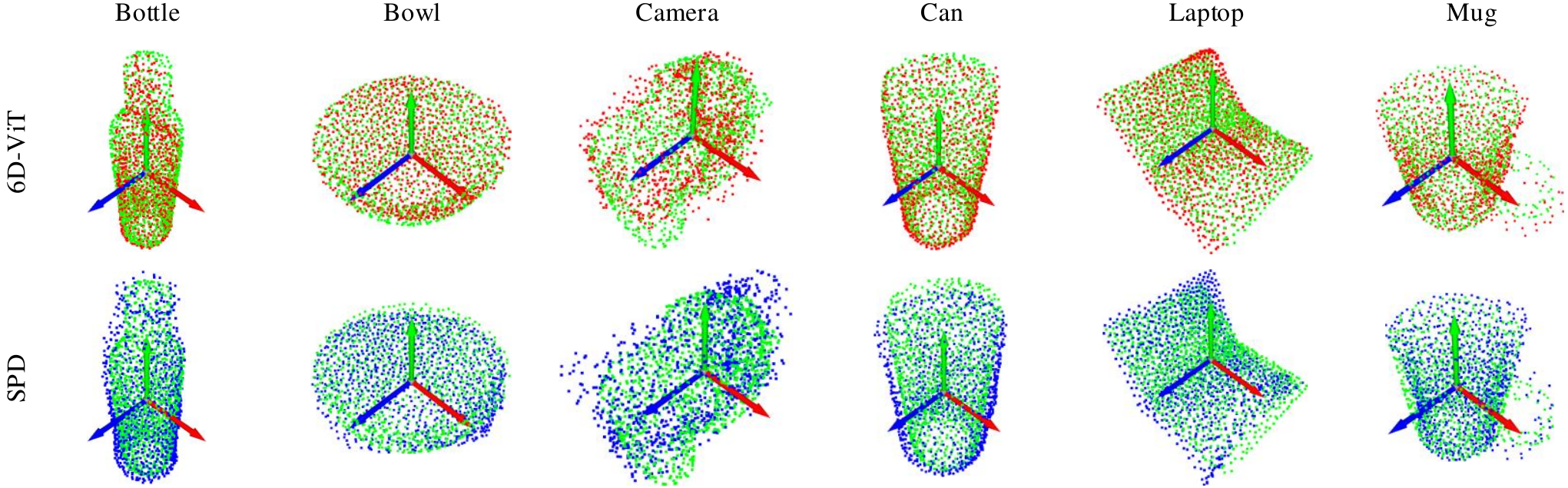}
\caption{Visualization of 3D object models reconstructed by our 6D-ViT and SPD~\cite{tian2020shape} on the CAMERA25 dataset. The red points represent our results, the blue points represent the SPD results, and the green points are the ground truth. We align all the results into the same pose.}
		\label{fig:fig_shape_camera}
\end{figure*}

\begin{figure*}[!ht]
		\centering
		\includegraphics[scale=0.85]{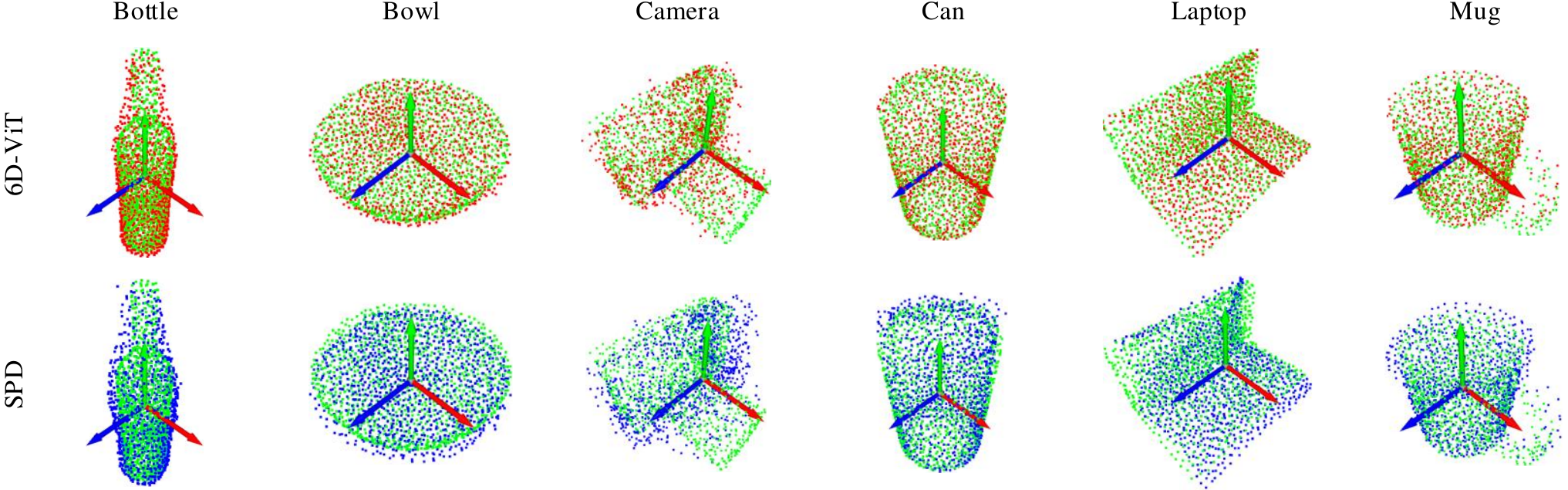}
\caption{Visualization of 3D object models reconstructed by our 6D-ViT and SPD~\cite{tian2020shape} on the REAL275 dataset. The red points represent our results, the blue points represent the SPD results, and the green points are the ground truth. We align all the visualization results into the same pose.}
		\label{fig:fig_shape_real}
\end{figure*}

We calculate the Chamfer distance (as defined in Eq.~\eqref{eq:cd}) between our reconstructed object models $\hat{\mathcal{P}}$ and the ground truth object models $\mathcal{P}$ in the canonical space to evaluate the quality of the model reconstruction. The results and comparisons with SPD~\cite{tian2020shape} are reported in \tablename~\ref{tab::shape}. It is observed that the 3D models reconstructed by our 6D-ViT obtain average CD metrics of 1.46 on the CAMERA25 dataset and 2.27 on the REAL275 dataset, compared to the 1.97 and 3.17, respectively, of the baseline. Better CD metrics indicate that the proposed instance representation learning framework improves the quality of the 3D model reconstruction. We also show a qualitative comparison between the 3D models reconstructed by our network and SPD~\cite{tian2020shape} for the CAMERA25 dataset in \figurename~\ref{fig:fig_shape_camera} and for the REAL275 dataset in \figurename~\ref{fig:fig_shape_real}.

From both the quantitative and qualitative perspectives, our 6D-ViT can steadily improve the reconstruction accuracy of the baseline~\cite{tian2020shape} over all categories on the CAMERA25 dataset and four out of six categories on the REAL275 dataset (the remaining two categories also achieve comparable performance). These analyses reveal that the instance representation learning framework we propose can not only improve the performance of 6D object pose estimation but also facilitate the reconstruction of canonical 3D object models.

\subsection{Ablation Analysis}
	\label{sec:ablation}

We conduct 18 ablation studies on the CAMERA25 and REAL275 datasets to investigate the efficiency of the individual components proposed in 6D-ViT. Specifically, we compare the proposed Pixelformer and Pointformer of our network with the latest state-of-the-art transformer-based appearance~\cite{wang2021pyramid} or geometric~\cite{zhao2021pointtransformer} feature generators as well as the feature extractors utilized in the baseline method~\cite{tian2020shape}. \tablename~\ref{tab::ablation_pair} shows different alternatives for the appearance and geometric feature generators. From the table, ``PSP", ``MLP", ``PVT", ``PT", ``PIF", and ``POF" refer to ``PSPNet in the baseline~\cite{tian2020shape}", ``linear projection layers in the baseline~\cite{tian2020shape}", ``pyramid vision transformer~\cite{wang2021pyramid}", ``point transformer~\cite{zhao2021pointtransformer}", ``Pixelformer in this work", and ``Pointformer in this work", respectively. There are nine combinations in total, and the corresponding comparisons are discussed below.

\subsubsection{Evaluation of Pixelformer}

To verify the effectiveness of our proposed Pixelformer (PIF), we compare it with two alternative RGB encoders: (1) PSPNet (PSP) with ResNet-18~\cite{he2016deep} as the backbone, which is consistent with the baseline method~\cite{tian2020shape}, and (2) PVT from~\cite{wang2021pyramid}, which like our PIF is a multiscale transformer structure designed for dense prediction tasks such as object detection and semantic segmentation. Note that there are a series of PVT models with different scales and that we employ PVT-tiny in the experiments since it shares the parameter settings listed in \tablename~\ref{tab::rgbformer} with ours.

As shown in \tablename~\ref{tab::ablation}, we set the geometric feature generators as MLPs in the baseline~\cite{tian2020shape}, the PT in~\cite{zhao2021pointtransformer} or our proposed POF and randomly select a backbone from the above encoders as the appearance feature extractor. The results are reported in \tablename~\ref{tab::ablation}. As shown in the table, comparing $\circled{1}$, $\circled{2}$ and $\circled{8}$; $\circled{5}$, $\circled{3}$ and $\circled{6}$; or $\circled{4}$, $\circled{7}$ and $\circled{9}$; we observe that regardless of whether the point cloud branch utilizes MLPs, the PT or our proposed POF, using our proposed PIF in the RGB branch achieves the best performance. This occurs on both datasets, which demonstrates the effectiveness of our Pixelformer. In addition, in all comparison experiments, our complete model ($\circled{9}$) is the best.

\subsubsection{Evaluation of Pointformer}

%$\circled{1}$ $\circled{5}$ $\circled{7}$ $\circled{2}$ $\circled{4}$ $\circled{3}$
Similarly, to verify the effectiveness of our proposed Pointformer (POF), we compare it with two alternative point cloud encoders: (1) MLPs with linear projection layers, which is consistent with the baseline~\cite{tian2020shape}, and (2) the PT in~\cite{zhao2021pointtransformer}, which is a multiscale transformer structure designed for point cloud processing. There are many variants of the PT, and we use the semantic segmentation structure, which demonstrates impressive performance gains in point cloud processing.

As shown in \tablename~\ref{tab::ablation}, we set the appearance feature generators as PSP~\cite{tian2020shape}, PVT~\cite{wang2021pyramid} or our proposed PIF and randomly select a backbone from the above encoders as the geometric feature extractors. The results are reported in \tablename~\ref{tab::ablation}. As shown in the table, comparing $\circled{1}$, $\circled{5}$ and $\circled{7}$; $\circled{2}$, $\circled{3}$ and $\circled{4}$; or $\circled{8}$, $\circled{6}$ and $\circled{9}$; we observe that regardless of whether the RGB branch utilizes PSP, PVT or our proposed PIF, using our proposed POF in the point cloud branch achieves the best performance. This occurs on both datasets, which demonstrates the effectiveness of our Pixelformer. In addition, in all comparison experiments, our complete model ($\circled{9}$) is the best.

\section{Conclusion}
	\label{sec:conclusion}

This paper presentes a novel instance representation learning framework for category-level 6D object pose estimation. To achieve this goal, two parallel transformer-based encoder-decoder networks are designed to learn the long-range dependent instance appearance and geometric characteristics from RGB images and point clouds. Then, the obtained instance features are fused with categorical shape features to generate joint and dense instance representations. Finally, the 6D object pose is obtained from post-alignment by solving the Umeyama algorithm. We conduct quantitative and qualitative evaluations on two public benchmark datasets to compare the presented method with the state-of-the-art methods. The results demonstrate that the proposed method achieves state-of-the-art performance on both datasets and performs significantly better than the existing methods. We also implement sufficient ablation experiments to prove the effectiveness of each component of the proposed network. In the future, we will focus on how to more efficiently explore the potential correlations between instance features from different sources and apply our instance representation learning framework to the instance-level pose estimation problem.

% or




\bibliographystyle{IEEEtran}
\bibliography{ref}


%\newpage


%\vfill


\end{document}